\documentclass[11pt]{article}
\PassOptionsToPackage{table}{xcolor}
\usepackage[preprint]{acl}

\usepackage{times}
\usepackage{latexsym}
\usepackage[T1]{fontenc}
\usepackage[utf8]{inputenc}
\usepackage{microtype}
\usepackage{inconsolata}
\usepackage{graphicx}

\usepackage{amsmath}
\usepackage{amssymb}
\usepackage{booktabs}
\usepackage{multirow}
\usepackage{array}
\usepackage{xspace}
\usepackage[inline]{enumitem}
\usepackage{tikz}
\usetikzlibrary{arrows.meta,positioning,calc,shapes.geometric,backgrounds}
\usepackage{pgfplots}
\pgfplotsset{compat=1.17}

\newif\ifshowtodos
\showtodostrue

\newcommand{\vibench}{ViBench\xspace}
\newcommand{\lead}{GLM~5.2\xspace}
\newcommand{\worker}{MiniMax~M3\xspace}

\newcommand{\tool}[1]{\texttt{#1}}

\newcommand{\Lm}{\mathcal{L}}            %
\newcommand{\Wm}{\mathcal{W}}            %
\newcommand{\Pset}{\mathcal{S}}          %
\newcommand{\Tset}{\mathcal{T}}          %
\newcommand{\gap}{\bar{g}}               %
\newcommand{\gapw}{g^{\max}}             %
\newcommand{\Sav}{\mathrm{Sav}}          %
\newcommand{\Risk}{\mathrm{Risk}}        %
\newcommand{\Dec}{D}                     %
\newcommand{\tierdel}{\textsc{delegate}\xspace}
\newcommand{\tiercond}{\textsc{conditional}\xspace}
\newcommand{\tierkeep}{\textsc{keep}\xspace}
\DeclareMathOperator*{\mean}{mean}

\definecolor{leadc}{HTML}{D1503F}
\definecolor{workerc}{HTML}{7759C2}
\definecolor{delc}{HTML}{2E9E6B}
\definecolor{condc}{HTML}{D4900B}
\definecolor{keepc}{HTML}{C44E3B}
\definecolor{armac}{HTML}{2A78D6}
\definecolor{oraclec}{HTML}{1F9BD1}
\definecolor{jevc}{HTML}{9B3F8F}
\definecolor{soloc}{HTML}{6B6B6B}
\definecolor{prc}{HTML}{1E8E5A}

\newcommand{\tD}{\cellcolor{delc!30}D}
\newcommand{\tC}{\cellcolor{condc!30}c}
\newcommand{\tCl}{\cellcolor{condc!30}c$^{*}$}
\newcommand{\tK}{\cellcolor{keepc!30}K}

\usepackage{fontawesome5} %

\newcommand{\sSchemaHO}{78.5}    \newcommand{\cSchemaHO}{\$0.67}
  
\newcommand{\sOracleHO}{71.0}    \newcommand{\cOracleHO}{\$0.79}  %
\newcommand{\sJevHO}{60.5}       \newcommand{\cJevHO}{\$0.72}
\newcommand{\savSchemaHO}{19\%}  %

\newcommand{\sProse}{87.9}       \newcommand{\cProse}{\$0.92}   %

\newcommand{\sSchemaSeedOne}{88.1} %

\newcommand{\nSteps}{1{,}055}

\newcommand{\nPhases}{9}
\newcommand{\segAgree}{98.7\%}
\newcommand{\swapDelta}{0.21}
\newcommand{\profileCost}{\$30}

\title{Partial Routing: Trace-Tuned Delegation for\\ Resource-Efficient Agentic Workflows}
\author{Siddharth Arya$^{1,2,*}$ \quad Kshitij Jain$^{2}$ \\
  $^{1}$University of Toronto \quad $^{2}$Georgian \\
  $^{*}$Correspondence: \texttt{siddharth.arya@georgian.io}}

\begin{document}
\maketitle
\raggedbottom
\begin{abstract}

AI models vary enormously in capability and price. Model routing exploits this by sending easier tasks to cheaper models. However, a single agentic task often combines easy and difficult subtasks, making one model choice for the entire task inefficient.
We introduce partial routing, which guides a frontier lead model to delegate suitable subtasks to a cheaper worker model. From both models' past runs, we identify recurring subtasks and measure each model's quality and cost on them. These measurements produce delegation guidance, with a parameter, $\delta$, controlling how much is delegated. The lead decides when and what to delegate during execution, without training either model.
On held-out tasks from ViBench and Terminal-Bench 4.0, partial routing achieves higher mean task performance than the lead alone and task-level routing. With $\delta = 0.75$, it raises the lead's ViBench score from 70.2 to 78.5 at 19\% lower cost, and its Terminal-Bench pass rate from 54\% to 61\% at similar cost.
Giving the lead only the subtask breakdown, without a worker, also reduces cost on both benchmarks and raises the Terminal-Bench pass rate. These results show that analysis of past runs can support more efficient planning and more effective delegation, with little cost and implementation overhead.

\end{abstract}

\section{Introduction}
\label{sec:intro}

Model routing assigns each request to the cheapest model likely to handle it, exploiting the fact that many requests need only a fraction of a frontier model's capability \citep{ong2025routellm,chen2023frugalgpt}. Both difficulty-estimation-based \citep{appliedcompute2026router} and cascade-based routing are deployed by practitioners \citep{georgian2026resource} for their  resource efficiency. Coding agents also now use routers to select optimal models for coding tasks based on measured context and complexity \citep{cursor2026router}.

Agents now complete tasks that span more and more hours of work \citep{kwa2025horizon}, and benchmarks follow this trend: SWE-Bench Pro collects changes that take engineers up to days \citep{deng2025swebenchpro}, and Terminal-Bench~4.0 allows runs of up to eight hours \citep{marten2026tb4}.
Within the larger job, an agent completes a sequence of subtasks that differ in difficulty. 
Assigning the entire task to a frontier model spends frontier prices on routine work; assigning it to a cheaper model exposes the difficult subtasks to that model's failures.

Coding subagent tools already let a lead agent delegate work during a rollout.
Claude Code, for example, can automatically spawn workers with separate contexts and configurable models \citep{anthropic2025subagents}.
Such \emph{unguided subagents} provide the mechanism for dynamic delegation, but what gets delegated, and as a consequence the cost and quality, is left entirely to the lead model's choice. 
\emph{Trained orchestrators} learn when to delegate from outcomes, at the price of training the lead and retraining it when models change (Section~\ref{sec:related}).

Effective delegation should be \emph{dynamic}, improvised as work emerges during the rollout, and \emph{informed}, based on measured model quality and cost on the target workload.
For production use it should also leave model weights untouched, so that it applies to proprietary frontier models and has less adoption overhead.

To address this, we introduce \emph{partial routing}, a method that tunes delegation from solo rollouts of the lead and worker models.
An LLM annotator identifies recurring subtasks in traces, and an LLM judge scores each model's work on each subtask.
Measured costs and quality gaps compile into a delegation policy, and a parameter $\delta$ gives control over how much work is delegated. 
The lead receives this policy in its prompt and chooses the scope and timing of each hand-off.
The lead's weights remain fixed. 
As Table~\ref{tab:positioning} shows, this makes our method of delegation uniquely dynamic, informed and controllable without training any model. 

We evaluate on two benchmarks built from real workloads (Appendix~\ref{app:industry}). In \vibench \citep{zhong2026vibench}, an agent builds a full-stack web application from a product requirements document (PRD) drawn from production traces on the Replit platform.\footnote{Replit is a Georgian portfolio company. Replit represents a subset of the firm's investments and may not be representative of the firm's investment experience or performance as a whole.} In Terminal-Bench~4.0 \citep{merrill2026terminalbench}, an agent completes command-line tasks inspired by real workflows. Partial routing can be deployed at two levels: the profiled subtasks as a plan for the lead alone, or the full delegation policy with a worker. We compare both with baselines that teams can deploy today, unguided delegation and an off-the-shelf router, and with a hindsight oracle router.  Unguided delegation gains 5.3 points over the lead alone on \vibench but none on Terminal-Bench; our plan alone (without worker) cuts the lead's cost by 39\% on \vibench and 16\% on Terminal-Bench at a similar or higher score. The full policy has the highest mean score on both benchmarks, 8.3 points above the lead alone on \vibench at \savSchemaHO{} lower cost.

Our contributions are (1)~a profiling method that attributes long agent rollouts to recurring subtasks, with quality scores and tiers that are stable across repeated runs; (2)~a policy that compiles subtask quality and cost into delegation guidance, so that parts of a task can be routed; and (3)~a controlled evaluation of guided delegation against solo models, task-level routers and unguided (out-of-the-box) delegation of frontier leads to cheaper workers.

\begin{table}[t]
\centering
\footnotesize
\setlength{\tabcolsep}{2pt}
\renewcommand{\arraystretch}{1.15}
\begin{tabular}{@{}>{\raggedright\arraybackslash}p{1.8cm} c c c >{\raggedright\arraybackslash}p{1.85cm}@{}}
\toprule
Approach & Dynamic & \shortstack{Informed/\\tuned} & \shortstack{Control-\\lable} & Tuning cost \\
\midrule
\multicolumn{5}{@{}l}{\emph{Task-level routing}} \\
Learned routers & -- & $\checkmark$ & $\checkmark$ & Router training \\
\midrule
\multicolumn{5}{@{}l}{\emph{Subtask routing}} \\
Model-card-based & -- & $\circ$ & $\checkmark$ & Benchmark profiles \\
Unguided subagents & $\checkmark$ & -- & -- & None \\
\midrule
\multicolumn{5}{@{}l}{\emph{Learning from trajectories}} \\
Trained orchestrators & $\checkmark$ & $\checkmark$ & -- & Training the lead \\
\midrule
\rowcolor{delc!10}
\textbf{Partial routing} & $\checkmark$ & $\checkmark$ & $\checkmark$ & \textbf{Trace profiling, no training} \\
\bottomrule
\end{tabular}
\caption{Approaches grouped as in Section~\ref{sec:related}. \emph{Dynamic}: delegation is decided during the rollout. \emph{Informed/tuned}: decisions draw on measured outcomes. \emph{Controllable}: the user sets how much work goes to the cheaper model. $\circ$: partly.}
\label{tab:positioning}
\end{table}

\section{Related Work}
\label{sec:related}

\paragraph{Task-level routing.}
RouteLLM learns model selection from preference data \citep{ong2025routellm}; FrugalGPT learns to cascade from weaker to stronger models based on model confidence \citep{chen2023frugalgpt}. \citet{moslem2026routing} survey many more task-level routers. 
Agent-as-a-Router brings this to coding with a fine-tuned agent that picks one model per task and stores verified outcomes in a memory \citep{zhou2026agentrouter}, and System One models such as Jev pick a model from short model descriptions \citep{typesafe2026jev}.
These \emph{learned task-level routers} mostly train on measured outcomes and expose a threshold or budget for control, but fix one model for the whole rollout; our oracle and Jev baselines represent them (Section~\ref{sec:baselines}).

\paragraph{Subtask routing before and during execution.}
Topaz assigns models to user-specified subtasks before execution, from profiles that map public benchmark scores onto general skills \citep{okamoto2026explainable}.
Such \emph{model-card-based subtask routers} are controllable, but draw on benchmark scores rather than measurements on the target workload, and their assignments stay fixed during execution.
Subagent tools \citep{anthropic2025subagents} and Recursive Language Models \citep{zhang2025rlm} split work during execution, but with these \emph{unguided subagents} the lead delegates on its own judgment, with no measured outcomes and little user control. 
Evaluations of such delegation have used same-model subagents \citep{mao2026delm}, designed delegation instructions \citep{ding2026mseval} or search and research workloads \citep{anthropic2026costintel}; our unguided baseline measures it with a cheaper worker on long coding tasks.

\paragraph{Learning delegation from trajectories.}
Past rollouts can inform delegation through a model's weights or its context.
\emph{Trained orchestrators} take the first route: Puppeteer uses a policy trained by reinforcement learning to select agents as the task evolves \citep{dang2025puppeteer}, Uno-Orchestra learns selective delegation with a cost-penalized reward \citep{cui2026unoorchestra}, and Fugu-Ultra learns to write a full execution workflow from the task description \citep{sakana2026fugu}.
They fix the cost--quality trade-off at training time and must be retrained when the models change, an upfront cost that inference-cost comparisons omit.
Methods such as ExpeL \citep{zhao2024expel}, GEPA \citep{agrawal2025gepa} and ACE \citep{zhang2025ace} take the second route, distilling traces into instructions, prompts or context, but do not apply it to model routing.
Partial routing takes this lightweight second route for delegation, at about \profileCost{} of profiling on \vibench (Appendix~\ref{app:profile}).

\section{Partial Routing}
\label{sec:method}

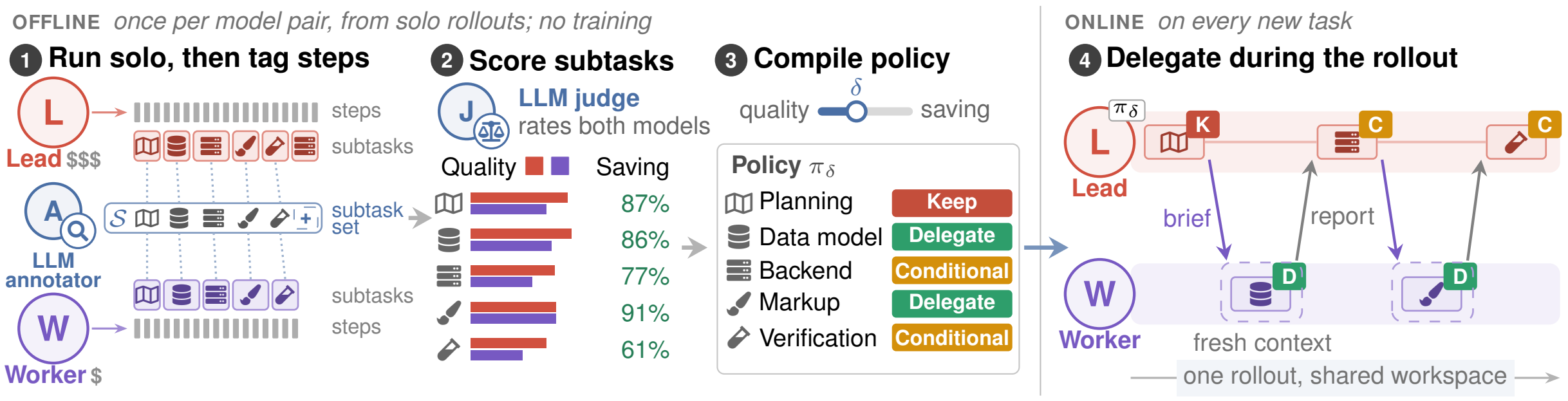
\begin{figure*}[t]
\centering
\begingroup
\definecolor{judgec}{HTML}{4A6FA5}
\newcommand{\fT}{\fontsize{8}{9}\selectfont}     %
\newcommand{\fL}{\fontsize{7}{8}\selectfont}     %
\newcommand{\fS}{\fontsize{6}{7}\selectfont}     %
\tikzset{
  zone/.style={rounded corners=4pt,fill=#1,draw=none},
  ptitle/.style={anchor=west,font=\sffamily\bfseries\fT,inner sep=0pt},
  zhead/.style={anchor=west,font=\sffamily\fL,text=black!55,inner sep=0pt},
  stepno/.style={circle,fill=black!75,text=white,font=\sffamily\bfseries\fS,inner sep=0pt,minimum size=3.3mm},
  lbl/.style={font=\sffamily\fL,inner sep=1pt},
  note/.style={font=\sffamily\fL,text=black!55,inner sep=1pt,align=center},
  aname/.style={font=\sffamily\bfseries\fL,text=#1,inner sep=0pt,anchor=north},
  flow/.style={-{Stealth[length=2mm,width=2.2mm]},line width=1.2pt,black!30},
  agent/.style={circle,draw=#1,line width=.8pt,fill=#1!8,text=#1,inner sep=0pt,minimum size=7mm,
    font=\sffamily\bfseries\fontsize{10}{10}\selectfont},
  helper/.style={agent=judgec,minimum size=5.6mm,font=\sffamily\bfseries\fontsize{8}{8}\selectfont},
  tool/.style={circle,fill=white,draw=judgec,line width=.6pt,text=judgec,font=\fS,inner sep=0pt,minimum size=3.5mm},
  span/.style={rounded corners=1.2pt,fill=#1!13,draw=#1!65,line width=.5pt,inner sep=0pt,
    minimum height=3.4mm,text=#1!75!black,font=\fS},
  tier/.style={rounded corners=1.2pt,fill=#1,text=white,font=\sffamily\bfseries\fS,inner sep=0pt,
    minimum width=3.4mm,minimum height=2.8mm},
  hand/.style={-{Stealth[length=1.6mm,width=1.6mm]},line width=.8pt,#1},
}
\newcommand{\icPlan}{\faMap[regular]}
\newcommand{\icDB}{\faDatabase}
\newcommand{\icAPI}{\faServer}
\newcommand{\icUI}{\faPaintBrush}
\newcommand{\icTest}{\faVial}
\resizebox{\textwidth}{!}{%
\begin{tikzpicture}[x=1cm,y=1cm]
\draw[black!25,line width=.4pt] (10.585,-.05) -- (10.585,-4.0);
\node[zhead] at (.15,-.22) {\textbf{\textsc{offline}}\enskip\textit{once per model pair, from solo rollouts; no training}};
\node[zhead] at (10.82,-.22) {\textbf{\textsc{online}}\enskip\textit{on every new task}};

\node[stepno] at (.3,-.6) {1};
\node[ptitle] at (.52,-.6) {Run solo, then tag steps};
\node[agent=leadc] (L1) at (.58,-1.13) {L};
\node[aname=leadc] at (.58,-1.5) {Lead\,\fS\textcolor{black!45}{\$\$\$}};
\draw[-{Stealth[length=1.3mm]},leadc!70,line width=.6pt] (.97,-1.13) -- (1.33,-1.13);
\foreach \i in {0,...,18} \fill[black!32,rounded corners=.3pt] ({1.4+\i*.1},-1.03) rectangle ++(.06,-.2);
\foreach \x/\w/\ic in {1.53/.26/\icPlan,1.83/.26/\icDB,2.18/.36/\icAPI,2.53/.26/\icUI,2.83/.26/\icTest,3.13/.26/\icAPI}
  \node[span=leadc,minimum width=\w cm,minimum height=3mm] at (\x,-1.47) {\ic};
\node[lbl,anchor=west,font=\sffamily\fS,text=black!50] at (3.36,-1.13) {steps};
\node[lbl,anchor=west,font=\sffamily\fS,text=black!50] at (3.36,-1.47) {subtasks};
\node[agent=workerc] (W1) at (.58,-3.32) {W};
\node[aname=workerc] at (.58,-3.69) {Worker\,\fS\textcolor{black!45}{\$}};
\draw[-{Stealth[length=1.3mm]},workerc!70,line width=.6pt] (.97,-3.32) -- (1.33,-3.32);
\foreach \i in {0,...,16} \fill[black!32,rounded corners=.3pt] ({1.4+\i*.1},-3.22) rectangle ++(.06,-.2);
\foreach \x/\w/\ic in {1.53/.26/\icPlan,1.88/.36/\icDB,2.23/.26/\icAPI,2.58/.36/\icUI,2.93/.26/\icTest}
  \node[span=workerc,minimum width=\w cm,minimum height=3mm] at (\x,-2.98) {\ic};
\node[lbl,anchor=west,font=\sffamily\fS,text=black!50] at (3.36,-3.32) {steps};
\node[lbl,anchor=west,font=\sffamily\fS,text=black!50] at (3.36,-2.98) {subtasks};
\node[helper] (ann) at (.58,-2.13) {A};
\node[tool] at ($(ann)+(.25,-.2)$) {\faSearch};
\node[anchor=north,inner sep=0pt,font=\sffamily\bfseries\fS,text=judgec,align=center] at (.58,-2.55) {LLM\\[-1.5pt]annotator};
\draw[rounded corners=2pt,fill=white,draw=judgec!70,line width=.5pt] (1.1,-2.05) rectangle (3.3,-2.35);
\node[font=\sffamily\bfseries\fL,text=judgec] at (1.24,-2.2) {$\Pset$};
\foreach \s/\l/\w/\ic in {1.53/1.53/1.53/\icPlan,1.855/1.83/1.88/\icDB,2.205/2.18/2.23/\icAPI,2.555/2.53/2.58/\icUI,2.88/2.83/2.93/\icTest}{
  \node[font=\fS,text=black!65] at (\s,-2.2) {\ic};
  \draw[judgec!55,densely dotted,line width=.6pt] (\s,-2.05) -- (\l,-1.63) (\s,-2.35) -- (\w,-2.82);
}
\node[draw=judgec!70,dashed,rounded corners=1pt,line width=.4pt,inner sep=0pt,minimum width=2.4mm,minimum height=2.2mm,
  font=\sffamily\bfseries\fS,text=judgec] at (3.14,-2.2) {+};
\node[lbl,anchor=west,font=\sffamily\fS,text=judgec,align=left] at (3.36,-2.2) {subtask\\[-2pt]set};

\draw[flow] (4.18,-2.2) -- (4.42,-2.2);

\begin{scope}[xshift=.45cm]
\node[stepno] at (4.12,-.6) {2};
\node[ptitle] at (4.34,-.6) {Score subtasks};
\node[helper] (jud) at (4.32,-1.12) {J};
\node[tool] at ($(jud)+(.25,-.2)$) {\faBalanceScale};
\node[lbl,anchor=west,align=left] at (4.8,-1.12) {\textcolor{judgec}{\textbf{LLM judge}}\\\textcolor{black!60}{rates both models}};
\node[lbl,anchor=west] at (4.02,-1.7) {Quality \textcolor{leadc}{$\blacksquare$}\,\textcolor{workerc}{$\blacksquare$}};
\node[lbl,anchor=east] at (6.42,-1.7) {Saving};
\foreach \y/\ic/\ql/\qw/\sv in {-2.05/\icPlan/8.2/6.4/87,-2.42/\icDB/8.5/6.8/86,-2.79/\icAPI/7.1/5.2/77,-3.16/\icUI/7.2/7.2/91,-3.53/\icTest/6.4/4.4/61}{
  \node[font=\fL,text=black!60] at (4.14,\y) {\ic};
  \fill[leadc] (4.36,\y+.11) rectangle ++(\ql*.12,-.1);
  \fill[workerc] (4.36,\y-.01) rectangle ++(\qw*.12,-.1);
  \node[lbl,anchor=east,text=delc!80!black] at (6.42,\y) {\sv\%};
}

\draw[flow] (6.5,-2.5) -- (6.76,-2.5);

\node[stepno] at (7.0,-.6) {3};
\node[ptitle] at (7.22,-.6) {Compile policy};
\node[lbl,anchor=east,text=black!60] at (7.84,-1.12) {quality};
\node[lbl,anchor=west,text=black!60] at (8.88,-1.12) {saving};
\draw[line width=2.4pt,black!15,line cap=round] (7.92,-1.12) -- (8.8,-1.12);
\draw[line width=2.4pt,judgec,line cap=round] (7.92,-1.12) -- (8.27,-1.12);
\filldraw[fill=white,draw=judgec,line width=1pt] (8.27,-1.12) circle (.095);
\node[lbl,anchor=south,text=judgec] at (8.27,-1.0) {$\delta$};
\draw[rounded corners=2pt,fill=white,draw=black!22] (6.86,-1.45) rectangle (9.92,-3.78);
\node[lbl,anchor=west,font=\sffamily\bfseries\fL,text=black!70] at (6.96,-1.68) {Policy $\pi_\delta$};
\foreach \y/\ic/\nm/\t/\tc in {-2.05/\icPlan/Planning/Keep/keepc,-2.39/\icDB/Data model/Delegate/delc,-2.73/\icAPI/Backend/Conditional/condc,-3.07/\icUI/Markup/Delegate/delc,-3.41/\icTest/Verification/Conditional/condc}{
  \node[font=\fL,text=black!60] at (7.08,\y) {\ic};
  \node[lbl,anchor=west] at (7.24,\y) {\nm};
  \node[tier=\tc,minimum width=1.22cm] at (9.24,\y) {\t};
}

\draw[flow,judgec!75] (9.97,-2.5) -- (10.42,-2.5);

\end{scope}

\begin{scope}[xshift=1.247cm,xscale=.922]
\node[stepno] at (10.62,-.6) {4};
\node[ptitle] at (10.84,-.6) {Delegate during the rollout};
\fill[leadc!9,rounded corners=3pt] (11.12,-1.12) rectangle (15.84,-1.74);
\fill[workerc!9,rounded corners=3pt] (11.12,-2.66) rectangle (15.84,-3.28);
\node[agent=leadc] (L) at (10.78,-1.43) {L};
\node[aname=leadc] at (10.78,-1.8) {Lead};
\node[draw=black!35,fill=white,rounded corners=1pt,inner sep=1pt,font=\fS] at ($(L)+(.3,.32)$) {$\pi_\delta$};
\node[agent=workerc] at (10.78,-2.97) {W};
\node[aname=workerc] at (10.78,-3.34) {Worker};
\foreach \n/\x/\ic/\t/\tc in {p/11.6/\icPlan/K/keepc,b/13.5/\icAPI/C/condc,v/15.36/\icTest/C/condc}{
  \node[span=leadc,minimum width=.6cm,font=\fL] (\n) at (\x,-1.43) {\ic};
  \node[tier=\tc] at ($(\n.north east)+(-.02,0)$) {\t};
}
\foreach \n/\x/\ic in {d/12.55/\icDB,u/14.43/\icUI}{
  \node[span=workerc,minimum width=.6cm,font=\fL] (\n) at (\x,-2.97) {\ic};
  \node[tier=delc] at ($(\n.north east)+(-.02,0)$) {D};
  \draw[workerc!60,dashed,rounded corners=3pt,line width=.5pt] ($(\n.south west)+(-.1,-.1)$) rectangle ($(\n.north east)+(.12,.14)$);
}
\draw[leadc!40,line width=.7pt] (p.east) -- (b.west) (b.east) -- (v.west);
\foreach \a/\w/\r in {p/d/b,b/u/v}{
  \draw[hand=workerc] ($(\a.south east)+(.05,0)$) -- ($(\w.north west)+(-.04,.16)$);
  \draw[hand=black!50] ($(\w.north east)+(.06,.16)$) -- ($(\r.south west)+(-.05,0)$);
}
\node[lbl,text=workerc,anchor=east] at (12.05,-2.2) {brief};
\node[lbl,text=black!55,anchor=west] at (13.05,-2.2) {report};
\node[note,anchor=north] at (12.57,-3.32) {fresh context};
\draw[-{Stealth[length=1.5mm]},black!30,line width=.6pt] (11.12,-3.82) -- node[note,fill=judgec!7,inner sep=2pt] {one rollout, shared workspace} (15.84,-3.82);
\end{scope}
\end{tikzpicture}%
}
\endgroup
\caption{Partial routing. Offline, the lead and the worker each do the same tasks alone, and an LLM annotator then tags each step of their rollouts (gray) with a recurring subtask from a set $\Pset$; an LLM judge scores both models on each subtask, and $\delta$ turns quality and cost saving into delegation tiers. Online, the lead follows this policy and briefs a fresh worker whenever it hands off a subtask. The five subtasks, their scores and their tiers at $\delta=0.40$ are from Table~\ref{tab:scorecard}.}
\label{fig:pipeline}
\end{figure*}

Figure~\ref{fig:pipeline} shows our complete pipeline, from solo traces to guidance for delegation during execution.

\subsection{Taxonomy: Tasks, subtasks and delegation}
\label{sec:setting}

An agent solves a task $x$ through a rollout $(e_1,\ldots,e_T)$ of messages and tool calls.
An evaluator, deterministic or LLM-based, assigns the completed task a score $r(x)\in[0,100]$.
$\Pset=\{s_1,\ldots,s_K\}$ is a set of recurring subtasks, such as planning, implementation and verification; each logged step $e_t$ receives one label $s_k$, and a subtask can recur throughout a rollout.

A \emph{delegation} is an assignment scoped dynamically by the lead model $\Lm$ and executed by the worker $\Wm$. The policy $\pi_\delta:\Pset\to\{\tierdel,\tiercond,\tierkeep\}$ guides the lead to delegate, decide case by case or retain each subtask.
A \tool{delegate\_subtask} tool starts a worker with the lead's tools, a shared workspace and a fresh context.
Each brief states the objective, relevant files, constraints and acceptance criteria.
The worker returns a report when it completes the assignment or reaches its step limit, a cap on the number of model calls it may make.

\subsection{Profiling solo rollouts}
\label{sec:profiling}

First, each model solves a set of tasks $\Tset$ alone. 
We retain the traces, token counts and final evaluation results.

\paragraph{Subtask annotation.}
An LLM annotator processes the solo rollouts $(e_1,\ldots,e_T)$ one at a time.
From the first rollout it processes, it seeds $\Pset$ with the subtasks it observes. 
For each subsequent rollout, it receives the trace and the current subtask set $\Pset$, with identifiers, names and definitions.
It assigns every step $e_t$ to a subtask $s_k$, pulling from $\Pset$ or adding a subtask when recurring work matches no existing subtask, updating $\Pset$ for the next rollout.
A final consolidation call merges duplicate subtasks and updates step assignments accordingly; no rollout is re-annotated.
This method ensures that our cost grows only linearly with additional rollouts, and it also allows practitioners to build a delegation policy from traces with sensitive user data that they cannot retain (Appendix~\ref{app:retention}).

\paragraph{Cost and quality.}
For model $m\in\{\Lm,\Wm\}$, subtask $s_k$ and task $i$, cost $c_{m,k,i}$ is the token cost of the attributed steps. 
Each step is charged for the single model request that produced it, priced from that request's measured uncached-input, cached-input and output tokens, so step costs sum exactly to the rollout's measured token totals.
We average $c_{m,k,i}$ over all tasks that contain that subtask to obtain $\bar c_{m,k}$.
An LLM judge sees the task specification, both models' attributed steps under anonymous A/B labels and their final test results, and returns a quality score $q_{m,k,i}\in[0,1]$ on a fixed rubric, with its confidence (Appendix~\ref{app:prompts}).
Annotation and judging are fully automated; Section~\ref{sec:profile-results} reports their stability over repeated runs.

\subsection{Compiling the delegation policy}
\label{sec:risk}

Each subtask now has cost and quality estimates for both the lead ($m=\Lm$) and the worker ($m=\Wm$). 
For each subtask, the estimated cost saving from delegating it is
\begin{equation}
\Sav_k=1-\frac{\bar c_{\Wm,k}}{\bar c_{\Lm,k}}.
\label{eq:savings}
\end{equation}
Quality risk combines the mean and worst-case quality drop when the worker replaces the lead.
Over tasks where both models performed the subtask, with $d_{k,i}=q_{\Lm,k,i}-q_{\Wm,k,i}$,
\begin{align}
\gap_k&=\max(0,\mean_i\,d_{k,i}),\\
\gapw_k&=\max(0,\max_i d_{k,i}),\\
\Risk_k&=(1-\rho)\gap_k+\rho\gapw_k.
\label{eq:risk}
\end{align}
The weight $\rho$ moves the penalty from the average drop ($\rho=0$) to the worst case ($\rho=1$), penalizing inconsistent quality; we fix $\rho=0.5$ and compute
\begin{equation}
\Dec_k(\delta)=\delta\Sav_k-(1-\delta)\Risk_k.
\label{eq:decision}
\end{equation}
Larger $\delta$ gives savings more weight, so more work is delegated.
A subtask is \tierdel when $\Dec_k\ge\tau$, \tierkeep when $\Dec_k\le-\tau$, and \tiercond otherwise; the threshold set at $\tau=0.15$, sets how many decisions are left to the lead during the rollout. A larger $\tau$ makes more subtasks \tiercond, while a smaller $\tau$ fixes more subtasks as \tierdel or \tierkeep.
Coordination subtasks such as planning and wrap-up, which the consolidation call marks as non-delegable, stay \tierkeep at every $\delta$, and low-confidence subtasks (Appendix~\ref{app:profile}) stay \tiercond.

The lead's prompt lists each subtask's identifier, one-line definition and tier (\emph{schema} guidance, Appendix~\ref{app:guidance}), fixed for the whole rollout.
A \emph{prose} variant that adds scores and advice from the judge's rationales is analyzed in Appendix~\ref{app:prose}.

\section{Experimental Setup}
\label{sec:setup}

\subsection{Benchmarks and splits}
\vibench \citep{zhong2026vibench} evaluates agents that build full-stack web applications from a PRD in an empty container; a browser-driving evaluation agent runs test plans against the result, and we report the mean plan score per application (0--100).
We profile on five applications and report results on ten held-out ones (profiling-set results in Appendix~\ref{app:train}).
On Terminal-Bench~4.0 \citep{merrill2026terminalbench,marten2026tb4}, an agent completes a command-line task in a container and passes if the task's pre-written tests pass; we profile on 12 tasks and report the pass rate on 23 held-out ones.
We use each benchmark's built-in automatic grading.

\subsection{Models, harness and accounting}
\label{sec:models}
On \vibench, the lead is \lead and the worker is \worker.
Their list input/output prices are \$1.40/\$4.40 and \$0.30/\$1.20 per million tokens, respectively, and cached input is charged at 10\% of the input price.
Both run in the \vibench OpenHands harness with a terminal, file editor and task tracker; routed arms add the delegation tool and share one base prompt.
We run the baselines (Section~\ref{sec:baselines}) and partial routing at $\delta\in\{0.10,0.40,0.75\}$ three times each to account for build and evaluation noise.  On Terminal-Bench, the lead is GPT-6.1 Sol and the worker GPT-6 Luna, at \$2/\$10 and \$0.10/\$0.50 per million input/output tokens respectively. 
They run in a separate, simpler harness with the same delegation tool, profiler, $\rho$ and $\tau$, and every method runs twice. Appendix~\ref{app:limits} lists the step and hand-off limits of every arm on both benchmarks.
Appendix~\ref{app:pair} describes a second model pair on each benchmark.

\subsection{Baselines}
\label{sec:baselines}

\paragraph{Solo and unguided baselines.}
Each solo model completes the task alone.
Unguided delegation gives the lead the delegation tool and base routing prompt but no profile; it is the primary control for the profile.

\paragraph{Task-level oracle router.}
With hindsight of each solo model's mean score per task, the oracle picks the cheaper model whose score clears a threshold $\theta\in\{0,10,\ldots,100\}$, and otherwise the higher-scoring one.
Tables report $\theta=100$, its highest-scoring setting, as a ceiling on task-level routing; Figure~\ref{fig:pareto} sweeps $\theta$.

\paragraph{Jev router.}
Jev \citep{typesafe2026jev}, a System One model built for fast, structured choices, reads the PRD (on Terminal-Bench, the task instruction) and a short card per model, and sends the task to the model it assigns the higher probability (Appendix~\ref{app:jev}).

\paragraph{Lead with a subtask plan.}

Partial routing can also be deployed without a worker: the lead works alone, with the profiled subtasks in its prompt as a proposed plan for the task, and no delegation tool or prompt (Appendix~\ref{app:plan}).
This baseline needs no quality judging and no worker at deployment. We also compare it to our full policy to isolate the uplift from routing alone.

\section{Results}
\label{sec:results}

\subsection{The learned subtask profiles}
\label{sec:profile-results}

On ViBench, the annotator induced \nPhases{} subtasks from the \nSteps{} steps of the ten solo profiling traces (Table~\ref{tab:scorecard}).
The worker is 61--91\% cheaper on every subtask, while quality gaps vary more across subtasks: the worker matches or beats the lead on markup and environment configuration but trails by about two points on backend logic and verification.

The tier columns show the delegation policy the lead model receives. Engineers can see before deployment which subtasks will be handed off at each $\delta$ and why. Additionally, since tiers are plain text in the lead's prompt, any of them can easily be overridden. A production team may want to keep certain sensitive logic such as authentication and session handling with the lead regardless of savings. 
Across four judge re-runs, quality scores vary by at most 0.3 points, far below the ${\sim}2$-point gaps between the models, and every re-run reproduces all tiers at $\delta=0.75$ (Appendix~\ref{app:stability}).
On Terminal-Bench, the annotator induced six subtasks from the 12 profiling tasks; averaged over subtasks, the worker scores 2.3 points lower than the lead and is 79\% cheaper (Table~\ref{tab:tb-scorecard}, Appendix~\ref{terminalbench}).

\begin{table}[t]
\centering
\newcommand{\sdv}[1]{{\scriptsize\,$\pm$#1}}
\small
\setlength{\tabcolsep}{2.5pt}
\renewcommand{\arraystretch}{1.08}
\begin{tabular}{@{}l ll r ccc@{}}
\toprule
& \multicolumn{2}{c}{Quality} & Saving & \multicolumn{3}{c}{Tier at $\delta =$} \\
\cmidrule(lr){2-3}\cmidrule(l){5-7}
Subtask & \multicolumn{1}{c}{$q_{\Lm}$} & \multicolumn{1}{c}{$q_{\Wm}$} & (\%) & .10 & .40 & .75 \\
\midrule
Planning$^{\ddagger}$ & 8.2\sdv{0.1} & 6.4\sdv{0.1} & 87 & \tK & \tK & \tK \\
Project scaffold & 7.6\sdv{0.1} & 6.0\sdv{0.2} & 84 & \tC & \tD & \tD \\
Data model/DB & 8.5\sdv{0.0} & 6.8\sdv{0.2} & 86 & \tK & \tD & \tD \\
Backend logic & 7.1\sdv{0.0} & 5.2\sdv{0.2} & 77 & \tK & \tC & \tD \\
Markup/styling & 7.2\sdv{0.0} & 7.2\sdv{0.1} & 91 & \tC & \tD & \tD \\
Client scripting & 5.8\sdv{0.1} & 4.2\sdv{0.1} & 77 & \tCl & \tCl & \tCl \\
Environment & 7.4\sdv{0.2} & 7.8\sdv{0.1} & 80 & \tC & \tD & \tD \\
Verification & 6.4\sdv{0.2} & 4.4\sdv{0.1} & 61 & \tK & \tC & \tD \\
Wrap-up$^{\ddagger}$ & 7.4\sdv{0.3} & 5.4\sdv{0.2} & 75 & \tK & \tK & \tK \\
\bottomrule
\end{tabular}
\caption{Subtask quality (0--10, $\pm$ SD over four judge re-runs), cost saving and tiers from five profiling applications. D = delegate, c = conditional, K = keep. $^{\ddagger}$Coordination. $^{*}$Low confidence. Appendices~\ref{app:profile} and~\ref{app:stability} give details.}
\label{tab:scorecard}
\end{table}

\subsection{Cost savings and quality increase}
\label{sec:main-results}

\begin{table}[t]
\centering
\small
\setlength{\tabcolsep}{5pt}
\renewcommand{\arraystretch}{1.1}
\begin{tabular}{@{}l rr rr@{}}
\toprule
& \multicolumn{2}{c}{ViBench} & \multicolumn{2}{c}{Terminal-Bench} \\
\cmidrule(lr){2-3}\cmidrule(l){4-5}
& Score\,$\uparrow$ & Cost\,$\downarrow$ & Pass\,$\uparrow$ & Cost\,$\downarrow$ \\
Method & (0--100) & (\$) & (\%) & (\$) \\
\midrule
\multicolumn{5}{@{}l}{\emph{Task-level routing}} \\
Lead alone & 70.2 & 0.83 & 54.3 & 1.17 \\
Worker alone & 46.9 & 0.20 & 8.7 & 0.39 \\
Jev & 60.5 & 0.72 & 50.0 & 1.05 \\
Oracle (hindsight) & 71.0 & 0.79 & 58.7 & 0.84 \\
Ours, plan only & 68.9 & 0.50 & 58.7 & 0.99 \\
\midrule
\multicolumn{5}{@{}l}{\emph{Subtask-level routing}} \\
Unguided & 75.6 & 0.59 & 54.3 & 1.07 \\
Ours, $\delta=0.10$ & 65.4 & 0.60 & 54.3 & 0.98 \\
Ours, $\delta=0.40$ & 72.9 & 0.66 & 54.3 & 1.08 \\
\rowcolor{delc!10}
Ours, $\delta=0.75$ & \textbf{78.5} & 0.67 & \textbf{60.9} & 1.19 \\
\bottomrule
\end{tabular}
\caption{Results on the ten held-out ViBench applications (mean of three runs) and 23 held-out Terminal-Bench tasks (mean of two runs). Costs are per build or task. Bold marks the best score and pass rate. Sections~\ref{sec:models} and~\ref{sec:baselines} define the models and baselines.}
\label{tab:main}
\end{table}

\begin{figure}[t]
\centering
\providecommand{\vbparetowidth}{\columnwidth}
\providecommand{\vbparetoheight}{6cm}
\begin{tikzpicture}
\tikzset{vb lab/.style={font=\scriptsize, text=black!80, inner sep=1.5pt}}
\begin{axis}[
  width=\vbparetowidth, height=\vbparetoheight,
  xmin=0.1, xmax=1.0, ymin=30, ymax=85,
  xtick={0.1,0.2,0.3,0.4,0.5,0.6,0.7,0.8,0.9,1.0}, ytick={30,40,50,60,70,80},
  xticklabel style={/pgf/number format/.cd, fixed, fixed zerofill, precision=1},
  xlabel={Cost per build (\$)}, ylabel={ViBench score},
  xlabel style={font=\footnotesize, yshift=2pt}, ylabel style={font=\footnotesize, yshift=-4pt},
  ticklabel style={font=\scriptsize},
  grid=major, grid style={black!8}, axis line style={black!45}, tick style={black!45},
  legend cell align=left,
  legend style={at={(0.985,0.03)}, anchor=south east, font=\scriptsize, draw=none, fill=none,
                row sep=-1pt, inner sep=2pt},
]
\addplot[black!40, densely dashed, thick, forget plot] coordinates {(0.198,46.87) (0.504,68.87) (0.590,75.56) (0.674,78.47)};
\addplot[only marks, mark=square*, mark size=2.3pt, soloc] coordinates {(0.830,70.21) (0.198,46.87)};
\addlegendentry{Single model}
\addplot[only marks, mark=square*, mark size=2.3pt, mark options={fill=white, draw=soloc, line width=0.8pt}] coordinates {(0.504,68.87)};
\addlegendentry{Lead + subtask plan}
\addplot[oraclec, thin, mark=o, mark size=1.1pt, forget plot]
  coordinates {(0.198,46.87) (0.323,49.32) (0.408,54.62) (0.611,57.46) (0.694,61.82) (0.734,65.15) (0.793,70.98)};
\addplot[only marks, mark=*, mark size=2.4pt, mark options={fill=white, draw=oraclec, line width=0.9pt}] coordinates {(0.793,70.98)};
\addlegendentry{Oracle router}
\addplot[only marks, mark=triangle*, mark size=3pt, jevc] coordinates {(0.721,60.53)};
\addlegendentry{Jev router}
\addplot[only marks, mark=*, mark size=2.6pt, mark options={fill=armac, draw=armac!70!black}] coordinates {(0.590,75.56)};
\addlegendentry{Unguided}
\addplot[only marks, mark=diamond*, mark size=3.4pt, mark options={fill=prc!55!black, draw=prc!55!black}] coordinates {(0.674,78.47)};
\addlegendentry{Partial routing}
\addplot[only marks, mark=diamond*, mark size=3.4pt, mark options={fill=prc!40, draw=prc!70!black, line width=0.6pt}, forget plot] coordinates {(0.657,72.89)};
\addplot[only marks, mark=diamond*, mark size=3.4pt, mark options={fill=white, draw=prc!70!black, line width=0.7pt}, forget plot] coordinates {(0.597,65.42)};
\node[vb lab, anchor=north, yshift=-3pt] at (axis cs:0.830,70.21) {\lead};
\node[vb lab, anchor=north west, xshift=2pt, yshift=-2pt] at (axis cs:0.198,46.87) {\worker};
\node[vb lab, anchor=south, yshift=3pt] at (axis cs:0.793,70.98) {Oracle};
\node[vb lab, anchor=west, xshift=4pt] at (axis cs:0.721,60.53) {Jev};
\node[vb lab, anchor=south east, xshift=-1pt, yshift=2pt] at (axis cs:0.504,68.87) {\lead + plan};
\node[vb lab, anchor=south east, xshift=1pt, yshift=2pt] at (axis cs:0.590,75.56) {Unguided};
\node[vb lab, anchor=north east, xshift=-3pt, yshift=-1pt] at (axis cs:0.597,65.42) {$\delta{=}0.10$};
\node[vb lab, anchor=north, xshift=2pt, yshift=-3pt] at (axis cs:0.657,72.89) {$\delta{=}0.40$};
\node[vb lab, anchor=west, xshift=5pt] at (axis cs:0.674,78.47) {$\delta{=}0.75$};
\end{axis}
\end{tikzpicture}
\caption{Cost and quality on the ten held-out \textbf{ViBench applications}, mean of three runs. The oracle curve sweeps $\theta$ with hindsight (marked point: $\theta=100$, as in Table~\ref{tab:main}); the dashed line is the Pareto frontier of methods without hindsight. Higher scores and lower costs are better.}
\label{fig:pareto}
\end{figure}
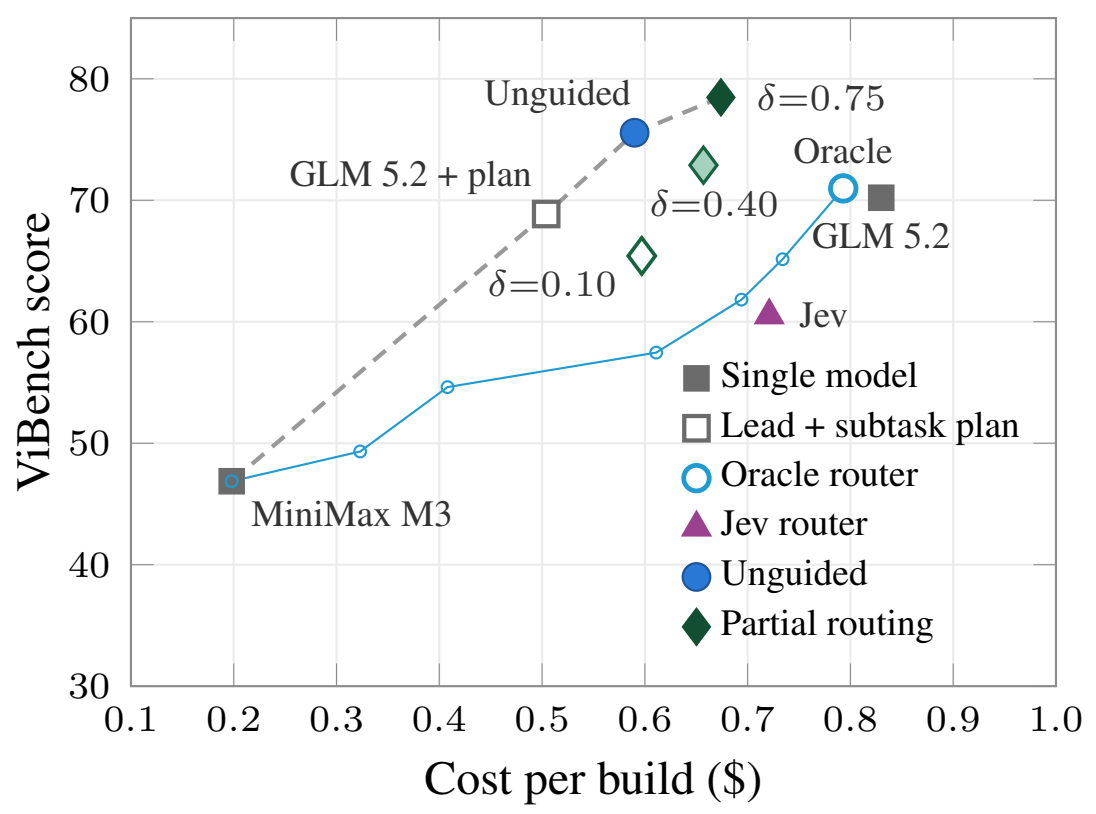

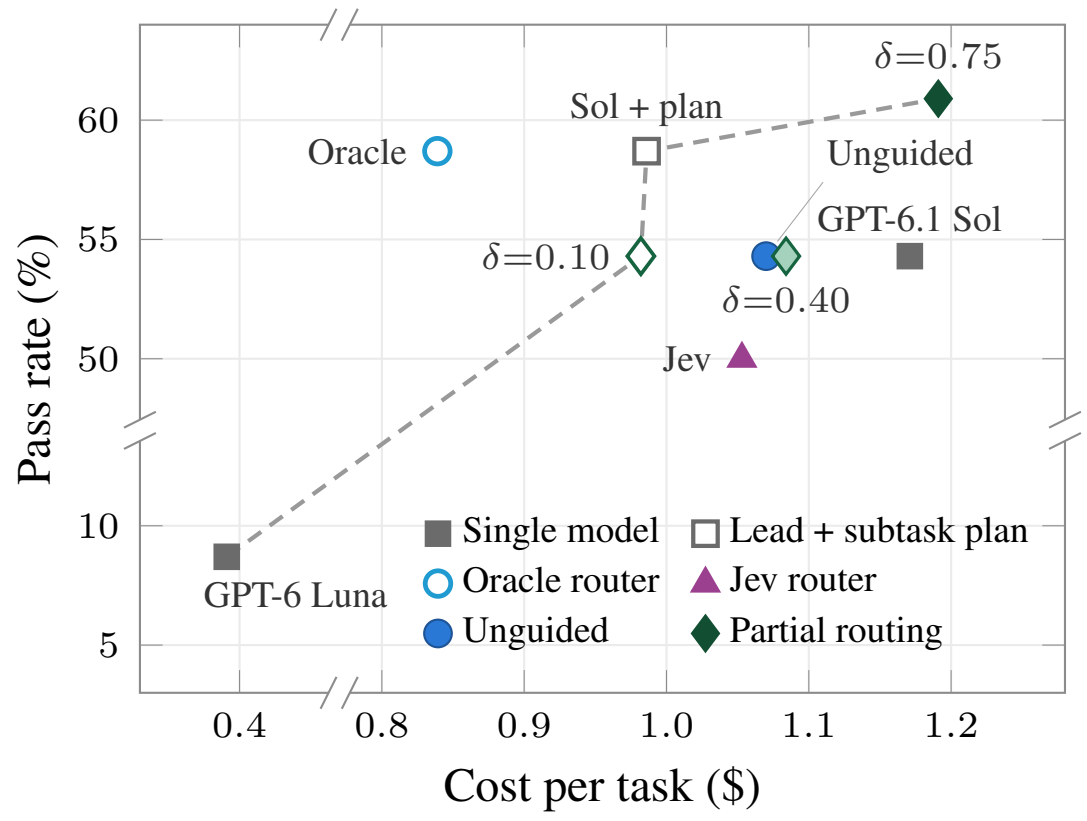
\begin{figure}[t]
\centering
\providecommand{\tbparetowidth}{\columnwidth}
\providecommand{\tbparetoheight}{6cm}
\begin{tikzpicture}
\tikzset{tb lab/.style={font=\scriptsize, text=black!80, inner sep=1.5pt},
         tb break/.style={black!45, line width=0.4pt}}
\begin{axis}[
  width=\tbparetowidth, height=\tbparetoheight,
  x coord trafo/.code={\pgfmathparse{#1 > 0.6 ? #1 - 0.30 : #1}},
  x coord inv trafo/.code={\pgfmathparse{#1 > 0.47 ? #1 + 0.30 : #1}},
  y coord trafo/.code={\pgfmathparse{#1 > 30 ? #1 - 33 : #1}},
  y coord inv trafo/.code={\pgfmathparse{#1 > 14 ? #1 + 33 : #1}},
  xmin=0.33, xmax=1.28, ymin=3, ymax=64,
  xtick={0.4,0.8,0.9,1.0,1.1,1.2}, ytick={5,10,50,55,60},
  xticklabel style={/pgf/number format/.cd, fixed, fixed zerofill, precision=1},
  xlabel={Cost per task (\$)}, ylabel={Pass rate (\%)},
  xlabel style={font=\footnotesize, yshift=2pt}, ylabel style={font=\footnotesize, yshift=-4pt},
  ticklabel style={font=\scriptsize},
  grid=major, grid style={black!8}, axis line style={black!45}, tick style={black!45},
  legend cell align=left, legend columns=2,
  legend style={at={(0.985,0.03)}, anchor=south east, font=\scriptsize, draw=none, fill=none,
                row sep=-1pt, inner sep=2pt, /tikz/every even column/.append style={column sep=4pt}},
]
\addplot[black!40, densely dashed, thick, forget plot] coordinates {(0.391,8.7) (0.982,54.3) (0.986,58.7) (1.191,60.9)};
\addplot[only marks, mark=square*, mark size=2.3pt, soloc] coordinates {(1.171,54.3) (0.391,8.7)};
\addlegendentry{Single model}
\addplot[only marks, mark=square*, mark size=2.3pt, mark options={fill=white, draw=soloc, line width=0.8pt}] coordinates {(0.986,58.7)};
\addlegendentry{Lead + subtask plan}
\addplot[only marks, mark=*, mark size=2.4pt, mark options={fill=white, draw=oraclec, line width=0.9pt}] coordinates {(0.839,58.7)};
\addlegendentry{Oracle router}
\addplot[only marks, mark=triangle*, mark size=3pt, jevc] coordinates {(1.053,50.0)};
\addlegendentry{Jev router}
\addplot[only marks, mark=*, mark size=2.6pt, mark options={fill=armac, draw=armac!70!black}] coordinates {(1.070,54.3)};
\addlegendentry{Unguided}
\addplot[only marks, mark=diamond*, mark size=3.4pt, mark options={fill=prc!55!black, draw=prc!55!black}] coordinates {(1.191,60.9)};
\addlegendentry{Partial routing}
\addplot[only marks, mark=diamond*, mark size=3.4pt, mark options={fill=prc!40, draw=prc!70!black, line width=0.6pt}, forget plot] coordinates {(1.084,54.3)};
\addplot[only marks, mark=diamond*, mark size=3.4pt, mark options={fill=white, draw=prc!70!black, line width=0.7pt}, forget plot] coordinates {(0.982,54.3)};
\node[tb lab, anchor=south, yshift=3pt] at (axis cs:1.171,54.3) {GPT-6.1 Sol};
\node[tb lab, anchor=north west, xshift=-6pt, yshift=-3pt] at (axis cs:0.391,8.7) {GPT-6 Luna};
\node[tb lab, anchor=south, yshift=3pt] at (axis cs:0.986,58.7) {Sol + plan};
\node[tb lab, anchor=east, xshift=-4pt] at (axis cs:0.839,58.7) {Oracle};
\node[tb lab, anchor=east, xshift=-4pt] at (axis cs:1.053,50.0) {Jev};
\draw[black!35, very thin, shorten <=3.5pt] (axis cs:1.070,54.3) -- (axis cs:1.110,57.4);
\node[tb lab, anchor=south west, xshift=-1pt] at (axis cs:1.110,57.4) {Unguided};
\node[tb lab, anchor=east, xshift=-4pt] at (axis cs:0.982,54.3) {$\delta{=}0.10$};
\node[tb lab, anchor=north, yshift=-4pt] at (axis cs:1.084,54.3) {$\delta{=}0.40$};
\node[tb lab, anchor=south, yshift=4pt] at (axis cs:1.191,60.9) {$\delta{=}0.75$};
\coordinate (xb) at (axis cs:0.47,14);
\coordinate (bl) at (rel axis cs:0,0);
\coordinate (tr) at (rel axis cs:1,1);
\end{axis}
\coordinate (xb1) at (xb |- bl); \coordinate (xb2) at (xb |- tr);
\coordinate (yb1) at (bl |- xb); \coordinate (yb2) at (tr |- xb);
\foreach \p in {xb1, xb2} {
  \fill[white] ([xshift=-2pt,yshift=-1pt]\p) rectangle ([xshift=2pt,yshift=1pt]\p);
  \draw[tb break] ([xshift=-3.5pt,yshift=-3pt]\p) -- ([xshift=-0.5pt,yshift=3pt]\p);
  \draw[tb break] ([xshift=0.5pt,yshift=-3pt]\p) -- ([xshift=3.5pt,yshift=3pt]\p);
}
\foreach \p in {yb1, yb2} {
  \fill[white] ([xshift=-1pt,yshift=-2pt]\p) rectangle ([xshift=1pt,yshift=2pt]\p);
  \draw[tb break] ([xshift=-3pt,yshift=-3.5pt]\p) -- ([xshift=3pt,yshift=-0.5pt]\p);
  \draw[tb break] ([xshift=-3pt,yshift=0.5pt]\p) -- ([xshift=3pt,yshift=3.5pt]\p);
}
\end{tikzpicture}
\caption{Cost and pass rate on the 23 held-out \textbf{Terminal-Bench tasks}, mean of two runs per task. Both axes are broken to fit GPT-6 Luna alone. Markers, labels and the dashed frontier are as in Figure~\ref{fig:pareto}.}
\label{fig:tb-pareto}
\end{figure}

On ViBench, partial routing at $\delta=0.75$ scores \sSchemaHO{} at \cSchemaHO{} per build, 8.3 points above the lead alone at \savSchemaHO{} lower cost (Table~\ref{tab:main}, Figure~\ref{fig:pareto}).
Unguided delegation already cuts the lead's cost by 29\% and gains 5.3 points; partial routing adds 2.9 points to this for \$0.08 more per build.
Both task-level routers score lower at higher cost: the hindsight oracle at most \sOracleHO{} at \cOracleHO{}, and Jev \sJevHO{} at \cJevHO{}.
Giving the lead the subtask list alone notably reduces cost while maintaining performance.

On Terminal-Bench, where the worker passes only 8.7\% of tasks alone, partial routing at $\delta=0.75$ passes 60.9\%, the most of any method, against 54.3\% for both the lead alone and unguided delegation, and 58.7\% for both the hindsight oracle and the lead with the plan (Figure~\ref{fig:tb-pareto}).
Here its gain is in quality rather than cost: it costs \$1.19 per task, about the lead's \$1.17. Surprisingly, giving the solo lead the subtask breakdown alone raises the pass rate while cutting cost.

We also ran partial routing at $\delta=0.75$ once with a second pair on each benchmark (Appendix~\ref{app:pair}).
On \vibench, with GPT-6.1 Sol leading GPT-6 Luna, it scores 88.2 at 37\% of Sol's cost, 5.2 points below Sol alone and 14.6 above Luna alone.
On Terminal-Bench, with Kimi K3.1 Pro leading DeepSeek V4.1 Flash, it passes 65.2\% of tasks, against 60.9\% for Kimi and 30.4\% for DeepSeek alone, at 8\% lower cost than Kimi.

\subsection{Effect of \texorpdfstring{$\delta$}{delta} on delegated work}
\label{sec:delta}

\begin{table}[t]
\centering
\small
\setlength{\tabcolsep}{2.5pt}
\renewcommand{\arraystretch}{1.1}
\begin{tabular}{@{}l r rrr@{}}
\toprule
& & \multicolumn{3}{c}{Ours, $\delta =$} \\
\cmidrule(l){3-5}
Measure & Unguided & .10 & .40 & .75 \\
\midrule
Hand-offs & 12 & 10 & 23 & 29 \\
Builds with worker & 10 & 10 & 18 & 19 \\
Worker turns & 315 & 292 & 568 & 793 \\
Turns/hand-off & 26.2 & 29.2 & 24.7 & 27.3 \\
\midrule
Turn share (\%) & 13.0 & 11.2 & 18.5 & 23.9 \\
Output share (\%) & 24.8 & 24.8 & 26.2 & 34.5 \\
Cost share (\%) & 4.7 & 4.9 & 6.0 & 8.0 \\
\bottomrule
\end{tabular}
\caption{Worker workload across the 30 held-out builds per method (ten applications, three runs). Shares divide worker totals by lead-plus-worker totals over all builds. Output tokens include commands and reports. Turns are model calls that use a tool.}
\label{tab:delegation}
\end{table}

Raising $\delta$ moves more work to the worker on both benchmarks (Tables~\ref{tab:delegation} and~\ref{tab:tb-delegation}).
On ViBench, from $\delta=0.10$ to $0.75$, hand-offs rise from 10 to 29 and the worker's share of turns from 11\% to 24\%; at $\delta=0.75$ workers generate a third of output tokens for 8\% of build cost.
On Terminal-Bench, at $\delta=0.10$ the lead keeps every subtask and never delegates, while at $\delta=0.75$ workers take 86\% of turns for 35\% of cost.
On both benchmarks, partial routing at $\delta=0.75$ delegates more work and scores higher than unguided delegation.

\section{Discussion and Conclusion}
\label{sec:conclusion}

\paragraph{Deploying partial routing.}
Partial routing needs only what production coding agents already have: subagent delegation \citep{anthropic2025subagents,replit2026agent4} and logged traces.
Our method adds a cheap offline profiling pass and a block of guidance in the lead's prompt.
No weights change, so it applies to proprietary API models, and the profile can be recompiled whenever a provider updates either model. The parameter $\delta$ can be changed without re-profiling, so we recommend sweeping it on a validation set of tasks to pick the best value.
Where retention policies forbid keeping user traces, the profile can be built as traces arrive and stored without them (Appendix~\ref{app:retention}).

\paragraph{Where does the quality gain come from?}
Partial routing gives the lead two things: a subtask plan and workers. Our baselines isolate the effect of each.

On held-out \vibench applications the workers account for most of the gain: unguided delegation scores 75.6 and the planned lead 68.9, against 70.2 for the lead alone, and partial routing at $\delta=0.75$ scores 9.6 points above the planned lead and 2.9 above unguided delegation.
We hypothesize that the workers' gain comes from context management: workers handle scoped assignments in fresh contexts and return compact reports, so the lead carries less implementation history, consistent with evidence that long inputs and distractors reduce model performance \citep{hong2025contextrot}.
On \vibench the plan mainly cuts cost, through fewer lead turns (Appendix~\ref{app:plan}).
On Terminal-Bench, by contrast, unguided delegation keeps the lead's pass rate at 9\% lower cost but gains no accuracy, while the plan accounts for part of the gain: the planned lead passes 58.7\% of tasks at 16\% lower cost.

\section*{Limitations}
Our main results use one model pair per benchmark, on ten held-out ViBench applications (three runs each) and 23 held-out Terminal-Bench tasks (two runs each).
The second pair on each benchmark ran once, against the two solo models only (Appendix~\ref{app:pair}); these runs indicate that the gains of partial routing carry over beyond the main pairs, but without the other baselines and repeated runs they are not a full evaluation of those pairs.
In every pair the worker is much cheaper and weaker than the lead, so whether the gains hold for pairs with smaller capability or price gaps is untested; our compute budget limited how many pairs, runs and baselines we ran, since each pair needs solo runs, baselines and every routed arm over several long rollouts.
Both benchmarks are coding and command-line workloads, chosen to represent long-running production agents (Appendix~\ref{app:industry}); evaluating partial routing on long-running agents in other domains, such as finance, healthcare and law, would show how well profiled subtasks and tiers transfer beyond software work.

Three settings were fixed throughout our experiments, and how sensitive the results are to each is still to be determined: the risk weight $\rho=0.5$, which balances the average and worst-case quality drop; the threshold $\tau=0.15$, which sets how many subtasks are left to the lead's judgment; and the number of profiling tasks (five ViBench applications and 12 Terminal-Bench tasks), which sets how much evidence each tier rests on.
A separate sweep of each would verify how much the tiers, and in turn the scores and costs, depend on it.

The profile's subtasks and quality scores come from an LLM annotator and judge; repeated runs show they are consistent (Section~\ref{sec:profile-results}), but we did not validate them against human labels.
Task outcomes, by contrast, come from each benchmark's own published grader, designed by its authors: test plans on ViBench and verification tests on Terminal-Bench (Appendix~\ref{app:industry}).

We did not systematically check whether the lead delegated exactly the subtasks its tiers assigned; how a lead follows, overrides or extends its tiers during a rollout is an interesting behavior to examine further.

Delegation, guided or not, trades wall-clock time for cost, since the lead waits for each worker's report: on Terminal-Bench the median task takes 69 minutes at $\delta=0.75$ and 55 with unguided delegation, against 14 for the lead alone (Table~\ref{tab:tb-delegation}).

\section*{Ethics Statement}
We use public benchmarks and API models, with generated applications executed in isolated containers. We used AI assistance to code the experiment harness and to edit the text; the authors checked all text, code, numbers, and references.

\clearpage
\appendix

\section{Profiler Checks and Cost}
\label{app:profile}

\paragraph{Models.}
The annotator and judge are Claude Fable 5.1 on \vibench and Claude Sonnet 5.5 on Terminal-Bench, called through the Claude Code command-line interface in headless mode with tools disabled and default generation settings; on \vibench, Claude Opus 5.5 wrote the one-line subtask definitions.
The annotator reads one application or task at a time in a fixed order, the lead's rollout before the worker's.
The stability re-runs (Appendix~\ref{app:stability}) and the second \vibench pair (Appendix~\ref{app:pair}) use Claude Fable 5.1 at high reasoning effort.
Profiling costs about \profileCost{} in LLM calls on \vibench.

\paragraph{Checks.}
The annotator's catch-all \emph{other} label covers 0.2\% of steps, and re-annotating one trace reproduces \segAgree{} of its labels. On one application, swapping which model the judge sees as A and which as B changes its 19 subtask quality scores by \swapDelta{} points on average.

\paragraph{Low confidence.}
A subtask is low-confidence if either model has fewer than two scored tasks, either quality-score standard deviation exceeds 2.5 on the 0--10 scale, or the judge reports low confidence.
On Terminal-Bench, whose pass/fail outcomes split the judge's scores, a subtask is low-confidence if the standard error of its paired lead--worker gap exceeds 1.5 points or the judge marks more than a third of its scores as low-confidence; applied to \vibench, this rule gives the same tiers at every $\delta$.

\paragraph{Full profile.}
Table~\ref{tab:scorecard-full} gives every cost, quality and risk term behind Table~\ref{tab:scorecard}, with tiers at six values of $\delta$.

\begin{table*}[t]
\centering
\normalsize
\setlength{\tabcolsep}{2.95pt}
\renewcommand{\arraystretch}{1.18}
\begin{tabular}{@{}l r rr r rr rrr cccccc@{}}
\toprule
& & \multicolumn{3}{c}{Cost per task} & \multicolumn{2}{c}{Quality} & \multicolumn{3}{c}{Risk terms} & \multicolumn{6}{c}{Tier at $\delta =$} \\
\cmidrule(lr){3-5}\cmidrule(lr){6-7}\cmidrule(lr){8-10}\cmidrule(l){11-16}
Subtask (induced) & Steps & $\bar c_{\Lm}$ & $\bar c_{\Wm}$ & $\Sav$ & $q_{\Lm}$ & $q_{\Wm}$ & $\gap$ & $\gapw$ & $\Risk$ & .10 & .20 & .30 & .40 & .50 & .75 \\
\midrule
Orientation \& planning$^{\ddagger}$ & 55  & .038 & .005 & .87 & 8.2 & 6.4 & .18 & .50 & .34 & \tK & \tK & \tK & \tK & \tK & \tK \\
Project scaffold                     & 58  & .054 & .009 & .84 & 7.6 & 6.0 & .16 & .30 & .23 & \tC & \tC & \tC & \tD & \tD & \tD \\
Data model \& DB setup               & 30  & .046 & .006 & .86 & 8.5 & 6.8 & .17 & .40 & .29 & \tK & \tC & \tC & \tD & \tD & \tD \\
Backend logic \& handlers            & 108 & .117 & .027 & .77 & 7.1 & 5.2 & .24 & .45 & .34 & \tK & \tC & \tC & \tC & \tD & \tD \\
Markup \& styling                    & 102 & .214 & .020 & .91 & 7.2 & 7.2 & .05 & .20 & .12 & \tC & \tC & \tD & \tD & \tD & \tD \\
Client-side scripting                & 54  & .106 & .025 & .77 & 5.8 & 4.2 & .15 & .60 & .38 & \tCl & \tCl & \tCl & \tCl & \tCl & \tCl \\
Environment configuration            & 92  & .072 & .014 & .80 & 7.4 & 7.8 & .00 & .10 & .05 & \tC & \tC & \tD & \tD & \tD & \tD \\
Verification                         & 533 & .384 & .150 & .61 & 6.4 & 4.4 & .20 & .40 & .30 & \tK & \tC & \tC & \tC & \tD & \tD \\
Wrap-up$^{\ddagger}$                 & 21  & .018 & .004 & .75 & 7.4 & 5.4 & .20 & .50 & .35 & \tK & \tK & \tK & \tK & \tK & \tK \\
\bottomrule
\end{tabular}
\caption{Full \vibench subtask profile. Costs are mean dollars per task containing the subtask; quality uses a 0--10 rubric. Mean and worst paired gaps are normalized by 10. Savings, risk and tiers are computed from unrounded values. D = delegate, c = conditional, K = keep. $^{*}$Low confidence; $^{\ddagger}$coordination retained by the lead.}
\label{tab:scorecard-full}
\end{table*}

\section{Profiler Prompts}
\label{app:prompts}

The prompts call subtasks \emph{phases} and the annotator a \emph{segmenter}.

\paragraph{Annotation.}
``Assign EVERY logged step to exactly one phase, reusing the phases already defined where they fit and defining new phases where nothing fits. [\ldots] Classify by what the step is FOR, not by which file it touches. [\ldots] Only define a new phase when the work is a genuinely different KIND of work that recurs (would appear in most tasks of this type), not a feature name, file name, or one-off.''

\paragraph{Quality rubric.}
9--10: complete and correct on the first pass, no rework, and downstream phases did not have to compensate.
7--8: works, with minor inefficiency, one rework loop or cosmetic issues.
5--6: mostly works, but with a real defect that a later phase or the evaluation caught, or noticeable thrash.
3--4: partially working, with defects that propagated downstream or were never fixed.
1--2: attempted but broken, or abandoned.
The judge is told not to reward or penalize speed or token usage, and to use evaluation results only where a test plausibly depends on the subtask.
We divide the judge's 0--10 score by 10 to obtain $q_{m,k,i}\in[0,1]$ (Section~\ref{sec:profiling}), so $\gap_k$, $\gapw_k$ and $\Risk_k$ lie in $[0,1]$; tables report quality on the 0--10 scale.

\section{Stability of the Profile}
\label{app:stability}

We repeated judging and annotation four times each on the ten solo traces of the five profiling applications (Table~\ref{tab:stability}).
Judging is highly repeatable: the judge re-runs, which score the original annotation, have an intraclass correlation of 0.97, every re-run reproduces the tiers at $\delta=0.75$, and at lower $\delta$ only project scaffold moves, by one tier, in some re-runs.
Annotation re-runs repeat annotation and consolidation from scratch and induce 8 or 9 subtasks, against 9 originally.
Matched to the re-run subtask that shares the most steps with it, an original subtask differs by a step-weighted Jaccard distance of 0.20, the same as between two re-runs; the largest distances are in client-side scripting (0.64) and wrap-up (0.54), neither of which is delegated by default.

\begin{table*}[t]
\centering
\newcommand{\sdv}[1]{{\footnotesize\,$\pm$#1}}
\normalsize
\setlength{\tabcolsep}{4pt}
\renewcommand{\arraystretch}{1.12}
\begin{tabular}{@{}l r ll lll l@{}}
\toprule
& & \multicolumn{5}{c}{Judge re-runs (same annotation)} & Annotation re-runs \\
\cmidrule(lr){3-7}\cmidrule(l){8-8}
Subtask & Steps & \multicolumn{1}{c}{$q_{\Lm}$} & \multicolumn{1}{c}{$q_{\Wm}$} & \multicolumn{1}{c}{$\gap$} & \multicolumn{1}{c}{$\gapw$} & \multicolumn{1}{c}{$\Risk$} & \multicolumn{1}{c}{Jaccard distance} \\
\midrule
Orientation \& planning$^{\ddagger}$ & 55  & 8.2\sdv{0.1} & 6.4\sdv{0.1} & .18\sdv{.01} & .50\sdv{.00} & .34\sdv{.01} & .22\sdv{.22} \\
Project scaffold                     & 58  & 7.6\sdv{0.1} & 6.0\sdv{0.2} & .16\sdv{.02} & .30\sdv{.08} & .23\sdv{.05} & .25\sdv{.09} \\
Data model \& DB setup               & 30  & 8.5\sdv{0.0} & 6.8\sdv{0.2} & .17\sdv{.02} & .40\sdv{.00} & .29\sdv{.01} & .23\sdv{.12} \\
Backend logic \& handlers            & 108 & 7.1\sdv{0.0} & 5.2\sdv{0.2} & .24\sdv{.02} & .45\sdv{.00} & .34\sdv{.01} & .18\sdv{.07} \\
Markup \& styling                    & 102 & 7.2\sdv{0.0} & 7.2\sdv{0.1} & .05\sdv{.01} & .20\sdv{.00} & .12\sdv{.01} & .27\sdv{.07} \\
Client-side scripting                & 54  & 5.8\sdv{0.1} & 4.2\sdv{0.1} & .15\sdv{.00} & .60\sdv{.00} & .38\sdv{.00} & .64\sdv{.24} \\
Environment configuration            & 92  & 7.4\sdv{0.2} & 7.8\sdv{0.1} & .00\sdv{.00} & .10\sdv{.00} & .05\sdv{.00} & .26\sdv{.05} \\
Verification                         & 533 & 6.4\sdv{0.2} & 4.4\sdv{0.1} & .20\sdv{.03} & .40\sdv{.10} & .30\sdv{.06} & .11\sdv{.03} \\
Wrap-up$^{\ddagger}$                 & 21  & 7.4\sdv{0.3} & 5.4\sdv{0.2} & .20\sdv{.04} & .50\sdv{.00} & .35\sdv{.02} & .54\sdv{.20} \\
\midrule
All steps                            & 1{,}053 & & & & & & .20\sdv{.04} \\
\bottomrule
\end{tabular}
\caption{Stability of the \vibench profile. Judge columns give the original value $\pm$ the standard deviation over four judge re-runs; quality is on the 0--10 scale, and gaps and risk are normalized by 10. Jaccard is the distance between a subtask's steps and those of its best-matching subtask in each of four annotation re-runs (mean $\pm$ SD; 0 means the same steps). $^{\ddagger}$Coordination, kept by the lead.}
\label{tab:stability}
\end{table*}

\section{Profiling under Data-Retention Restrictions}
\label{app:retention}

The delegation policy depends on the traces only through per-task costs and quality scores, so it can be profiled where retention policies forbid keeping user traces.
Each task $x$ is processed once both models' solo rollouts of it are available:
\begin{enumerate}[itemsep=1pt]
\item The annotator labels both rollouts against the current subtask set $\Pset$, adding subtasks as in Section~\ref{sec:profiling}.
\item The judge scores both models on every subtask labeled in $x$, and the attributed token counts give their costs.
\item The system stores each cost $c_{m,k,x}$, quality score $q_{m,k,x}$ and the judge's confidence, with the updated $\Pset$; the raw rollouts of $x$ can then be deleted.
\end{enumerate}
When consolidation merges two subtasks that occur in the same task, their costs are added and their scores averaged.
Equations~\ref{eq:savings}--\ref{eq:decision} and the low-confidence rules (Appendix~\ref{app:profile}) use only these records, so the policy can be compiled, and recompiled at a new $\delta$, without the traces; only the checks of Appendix~\ref{app:profile} need the traces, within the retention window.

\section{Guidance Examples}
\label{app:guidance}

All examples are from \vibench.

\paragraph{Base routing prompt (all routed arms).}
Every routed arm, including unguided delegation, receives the same block.
It tells the lead that the worker ``runs on a smaller, cheaper model'' with the same tools and workspace and its own copy of the PRD.
It says that ``delegating is how this build saves cost; verifying is how it stays correct,'' and asks for short briefs (objective, files, contracts the worker cannot infer, acceptance criteria and boundaries).
It states the limits of 6 delegations and 35 worker steps, and instructs the lead to verify every return itself rather than trust the worker's report.

\paragraph{Schema guidance, $\delta=0.75$ (abridged).}
The prompts call subtasks \emph{phases}.
\begin{quote}\small\ttfamily\raggedright
Each phase of the build below has a tier that says whether that kind of work goes to the worker or stays with you.\\
tiers:\\
\hspace*{1em}delegate: hand this work to the worker by default.\\
\hspace*{1em}conditional: you decide case by case whether to hand this work to the worker or do it yourself.\\
\hspace*{1em}keep: do this work yourself; do not hand it to the worker.\\
phases:\\
\hspace*{1em}- id: orientation, tier: keep\\
\hspace*{1em}- id: project\_scaffold, tier: delegate\\
\hspace*{3em}description: Creating the empty project skeleton: directory layout, dependency manifests, entry point, framework settings and extension wiring, without features.\\
\hspace*{1em}- id: templating, tier: delegate\\
\hspace*{1em}- id: client\_scripting, tier: conditional\\
\hspace*{1em}- id: verification, tier: delegate\\
\hspace*{1em}[\ldots data\_model, routing, environment\_config: delegate; wrap\_up: keep]
\end{quote}

\paragraph{Prose guidance, $\delta=0.75$ (excerpt).}
\begin{quote}\small
Cost context: the worker is roughly 3.7x cheaper than the lead per build (measured on the training builds).
You are the lead on this build and you have 6 delegations available. The default is to hand off: every unit of work listed under ``hand these off'' below should go to the worker unless it is blocked on a decision only you can make, and you are expected to use most of your delegations. [\ldots]

\textbf{Work the worker handles well --- hand these off by default}\\
\textbf{Markup \& styling}: the worker scored 7.2/10 (4 to 9) vs the lead 7.2/10 (6 to 9) across 4 builds, at 0.10$\times$ the cost the lead spends on this work. This is the safest hand-off in the build: the worker produces complete shells, stylesheets, test ids, ARIA attributes and empty states in a single pass [\ldots]. The recurring failure is template thrash when server templates share global helpers or extend one another inconsistently [\ldots]. The lead's check is simply to render every template once with real data and confirm the spec's required elements, ids and error blocks are present.

\textbf{Work that varies --- hand off when the piece is self-contained, otherwise keep it}\\
\textbf{Client-side scripting}: the worker scored 4.2/10 (3 to 6) vs the lead 5.8/10 (1 to 9) across 4 builds, at 0.23$\times$ the cost [\ldots]. Hand off when the script is a single controller over a server-rendered page [\ldots]. Keep it when the spec depends on the client for a correctness rule such as per-tab session scoping, offline reload via a service worker [\ldots].
\end{quote}

\section{Prose Guidance and Verification Behavior}
\label{app:prose}

Prose guidance adds the judge's scores and advice composed from its rationales (Appendix~\ref{app:guidance}).
At $\delta=0.75$ it tells the lead to exercise the specification's core flows in a browser, so it changes how the lead verifies as well as what it delegates: in its one run, browser use rises to 9 of 15 builds, against 2 under unguided delegation and 1 under schema guidance (Table~\ref{tab:qa}).
On that run schema guidance scores \sSchemaSeedOne{} at \$0.60 and prose guidance \sProse{} at \cProse{}, so schema guidance, which the main results use, reaches the same score at two-thirds of the cost without verification advice.

\begin{table}[h]
\centering
\scriptsize
\setlength{\tabcolsep}{2.5pt}
\begin{tabular}{@{}l r rr r rr@{}}
\toprule
& Hand- & \multicolumn{2}{c}{Lead steps} & Browser & \multicolumn{2}{c}{Run 1} \\
\cmidrule(lr){3-4}\cmidrule(l){6-7}
Method & offs & all & verif. & QA & Score & \$ \\
\midrule
Lead alone                    & --  & --  & --   & 3/15 & 74.3 & 1.07 \\
Unguided                      & 10  & 71  & 35.6 & 2/15 & 73.3 & 0.64 \\
$\delta{=}0.40$, prose        & 15  & 122 & 86.5 & 8/15 & 80.8 & 1.06 \\
$\delta{=}0.40$, schema       & 15  & 88  & 47.7 & 5/15 & 85.6 & 0.82 \\
$\delta{=}0.75$, prose        & 21  & 100 & 60.3 & 9/15 & 87.9 & 0.92 \\
$\delta{=}0.75$, schema       & 10  & 79  & 43.0 & 1/15 & 88.1 & 0.60 \\
\bottomrule
\end{tabular}
\caption{Lead behavior and results on run 1, over 15 applications. Hand-offs are totals over the 15 builds; lead steps are per-build means. Verification steps are a subset of lead steps. Browser QA counts builds where the lead drove a headless browser.}
\label{tab:qa}
\end{table}

\section{Benchmarks as Production Workloads}
\label{app:industry}

\vibench's PRDs are drawn from anonymized Replit production traces, and its evaluation agent drives each application in a browser against natural-language test plans, matching the production criterion that ``the app should work when users click around'' \citep{replit2026closingloop}; Replit uses \vibench alongside online A/B tests to vet changes to its production agent.
Terminal-Bench tasks are inspired by real command-line workflows, each with a human-written solution and verification tests \citep{merrill2026terminalbench}, and version~4.0 allows runs of up to eight hours \citep{marten2026tb4}.

\section{Run Limits}
\label{app:limits}

Table~\ref{tab:limits} lists the limits of every arm on both benchmarks.
A step is one model call, whether or not it uses a tool.
The lead and solo limits apply to every solo run, including the profiling runs and the lead with a subtask plan, and to the lead in every routed arm.

\begin{table}[h]
\centering
\small
\setlength{\tabcolsep}{3pt}
\begin{tabular}{@{}p{0.46\columnwidth}p{0.22\columnwidth}p{0.26\columnwidth}@{}}
\toprule
Limit & ViBench & Terminal-Bench \\
\midrule
Lead or solo model, steps per build or task & 300 & 300 \\
Worker, steps per hand-off & 35 & 120 \\
Hand-offs per build or task & 6 & 8 \\
\midrule
Lead or solo model told its own step limit & yes & no \\
Lead told the hand-off and worker limits & yes & yes \\
Worker told its step limit & yes & yes \\
Running count of steps used & every 10 steps & none \\
\bottomrule
\end{tabular}
\caption{Step and hand-off limits per arm. Worker steps do not count toward the lead's limit. A run that reaches the lead or solo limit is graded on the state it leaves; a worker that reaches its limit returns its report to the lead; once all hand-offs are used, the lead does the remaining work itself. No lead or solo run on the held-out tasks reached its step limit.}
\label{tab:limits}
\end{table}

\section{Jev Router}
\label{app:jev}

The Jev baseline (Section~\ref{sec:baselines}) calls \texttt{typesafe/jev-1.13} (served as \texttt{typesafe/jev-1.13-20260917}) through OpenRouter's decisions API.
Each request holds a fixed task sentence, the task input (the PRD on ViBench, the task instruction on Terminal-Bench), fixed instructions and one card per candidate model; nothing from the profiler, the traces or the solo outcomes enters it.
Each application or task is queried in both option orders, three times each, and the lead is picked when its mean probability over the six calls is at least 0.5.
Reported costs include two calls per pick, one per option order, which add about \$0.0002 per ViBench application and \$0.0001 per Terminal-Bench task; counting all six calls leaves every reported cost unchanged at two decimals.

\paragraph{ViBench request.}
\begin{quote}\small
\textbf{Task:} A coding agent will build the web application described in \texttt{prd}, from an empty workspace, working alone from start to finish. The finished app is graded by automated QA test plans that exercise it in a browser.\\
\textbf{Instructions:} Which model should run the coding agent that builds this app? Balance the quality of the finished app against cost: choose the cheaper model when it is likely to deliver a working, complete app, and the stronger model when the app is demanding enough that the cheaper model would likely leave it broken or incomplete.\\
\textbf{GLM 5.2:} GLM 5.2 (Z.ai): a frontier-class coding model, stronger on complex, multi-part builds. List price about \$1.40 per million input tokens and \$4.40 per million output tokens, roughly four times the cost of MiniMax M3.\\
\textbf{MiniMax M3:} MiniMax M3: a fast, low-cost model that handles routine coding well. List price about \$0.30 per million input tokens and \$1.20 per million output tokens.
\end{quote}

\paragraph{Terminal-Bench request.}
\begin{quote}\small
\textbf{Task:} A coding agent will complete the terminal task described in \texttt{instruction}, inside a Linux container with shell access, working alone from start to finish. The finished work is graded by automated tests run against the container's final state.\\
\textbf{Instructions:} Which model should run the coding agent that completes this task? Balance the quality of the finished work against cost: choose the cheaper model when it is likely to complete the task correctly, and the stronger model when the task is demanding enough that the cheaper model would likely leave it failing or incomplete.\\
\textbf{GPT-6.1 Sol:} GPT-6.1 Sol (OpenAI): a frontier-class coding model, stronger on complex, multi-step tasks. List price about \$2 per million input tokens and \$10 per million output tokens, roughly twenty times the cost of GPT-6 Luna.\\
\textbf{GPT-6 Luna:} GPT-6 Luna (OpenAI): a fast, low-cost model that handles routine coding well. List price about \$0.10 per million input tokens and \$0.50 per million output tokens.
\end{quote}

\section{Lead with a Subtask Plan}
\label{app:plan}

This baseline (Section~\ref{sec:baselines}) runs the lead alone with the profiled subtasks in its prompt as a proposed plan, without the worker, the delegation tool, the routing prompt or the tiers.
Its subtask records are those of the schema guidance with the tier lines removed (Appendix~\ref{app:guidance}), and each benchmark has one plan, used for every task and run.
The ViBench block, verbatim:
\begin{quote}\small\ttfamily\raggedright
<BUILD\_PLAN>\\
Use this as a learned prior for how the build breaks down. It is derived from previous traces of similar builds. Prefer it when splitting your work into tasks, but still obey the PRD and current repository state.\\[3pt]
Your build will involve the following phases of work.\\[3pt]
phases:
\begin{itemize}[leftmargin=2em,itemsep=1pt,topsep=1pt,parsep=0pt,label=-]
\item id: orientation\\ name: Orientation \& planning\\ description: Reading the product spec, inspecting the workspace and runtime environment, and drafting or revising the task plan before the closing update.
\item id: project\_scaffold\\ name: Project scaffold\\ description: Creating the empty project skeleton: directory layout, dependency manifests, entry point, framework settings and extension wiring, without features.
\item id: data\_model\\ name: Data model \& DB setup\\ description: Defining or changing the persistence layer: schemas, ORM models, constraints, and migration, table-creation or seed-data scripts.
\item id: routing\\ name: Backend logic \& handlers\\ description: Writing or fixing server-side code for requests and business rules: routes, views, forms, endpoints and their helper functions.
\item id: templating\\ name: Markup \& styling\\ description: Writing or editing HTML markup, server templates, static page shells, stylesheets, and static visual assets such as icons or images.
\item id: client\_scripting\\ name: Client-side scripting\\ description: Writing or fixing browser-side scripts for interactivity, client state, API calls, client storage, service workers and app manifests.
\item id: environment\_config\\ name: Environment configuration\\ description: Writing setup or start scripts, installing runtime dependencies, setting environment variables, and fixing errors that stop the application booting.
\item id: verification\\ name: Verification\\ description: Running an already-bootable application and exercising it through requests, database queries, logs, code searches or tests, including related setup and cleanup.
\item id: wrap\_up\\ name: Wrap-up\\ description: The final task-tracker update, listing of deliverables, removal of temporary files, and the completion summary ending the run.
\end{itemize}
</BUILD\_PLAN>
\end{quote}
On Terminal-Bench the block is \texttt{<TASK\_PLAN>}, worded for a task in a container, with the six subtasks of Table~\ref{tab:tb-scorecard}.

\paragraph{Runs.}
The planned lead uses the settings and step limit of the lead alone (Table~\ref{tab:limits}) and runs three times on \vibench and twice on Terminal-Bench, like every method.

\paragraph{Results.}
On the held-out \vibench applications the planned lead takes 74 lead turns per build against 100 for the lead alone and costs 39\% less (\$0.50 against \$0.83), while fully passing as many test plans as the lead alone (46 of 87: 29 plans in each of three runs).
It scores 1.3 points below the lead alone, and partial routing at $\delta=0.75$ scores 9.6 points above it.
On Terminal-Bench it passes 13.5 of 23 tasks on average at \$0.99 per task.
On both benchmarks it lies on the cost--quality frontier (Figures~\ref{fig:pareto} and~\ref{fig:tb-pareto}).

\section{Results on the Profiling Applications}
\label{app:train}

The main text reports \vibench on the ten held-out applications.
Table~\ref{tab:train} adds the five profiling applications, whose solo traces produced the profile and whose test results the judge saw (Section~\ref{sec:profiling}).

\begin{table*}[t]
\centering
\small
\setlength{\tabcolsep}{5pt}
\renewcommand{\arraystretch}{1.1}
\begin{tabular}{@{}l rr rr rr@{}}
\toprule
& \multicolumn{2}{c}{Profiling (5)} & \multicolumn{2}{c}{Held-out (10)} & \multicolumn{2}{c}{All (15)} \\
\cmidrule(lr){2-3}\cmidrule(lr){4-5}\cmidrule(l){6-7}
Method & Score & \$ & Score & \$ & Score & \$ \\
\midrule
\multicolumn{7}{@{}l}{\emph{Task-level routing}} \\
\lead alone & 71.8 & 0.71 & 70.2 & 0.83 & 70.7 & 0.79 \\
\worker alone & 44.8 & 0.19 & 46.9 & 0.20 & 46.2 & 0.20 \\
Jev & 57.8 & 0.47 & 60.5 & 0.72 & 59.6 & 0.64 \\
Oracle (hindsight) & 74.4 & 0.58 & 71.0 & 0.79 & 72.1 & 0.72 \\
Ours, plan only & 59.7 & 0.65 & 68.9 & 0.50 & 65.8 & 0.55 \\
\midrule
\multicolumn{7}{@{}l}{\emph{Subtask-level routing}} \\
Unguided & 70.3 & 0.43 & 75.6 & 0.59 & 73.8 & 0.54 \\
Ours, $\delta=0.10$ & 82.5 & 0.46 & 65.4 & 0.60 & 71.1 & 0.55 \\
Ours, $\delta=0.40$ & \textbf{86.0} & 0.56 & 72.9 & 0.66 & 77.3 & 0.63 \\
\rowcolor{delc!10}
Ours, $\delta=0.75$ & 85.5 & 0.43 & \textbf{78.5} & 0.67 & \textbf{80.8} & 0.59 \\
\bottomrule
\end{tabular}
\caption{ViBench score (0--100) and cost (\$ per build), averaged over three runs per application. Profiling comprises the five profiling applications; held-out is the set of Table~\ref{tab:main}. Bold marks the highest score in each column. Ours uses schema guidance. The oracle ($\theta=100$) selects, with hindsight, the solo model with the higher three-run mean score on each application.}
\label{tab:train}
\end{table*}

The guided arms score higher on the profiling applications than on the held-out ones, by 7.0 points at $\delta=0.75$ and 13--17 points at $\delta=0.10$ and $0.40$, while unguided delegation scores 5.2 points lower and the lead alone 1.6 points higher.
At $\delta=0.75$ partial routing leads unguided delegation by 15.2 points on the profiling applications and 2.9 on the held-out ones.
The subtask list alone does not help on the applications it came from: the lead with a subtask plan scores 12.1 points below the lead alone there.
Because the profile comes from these applications, the main text evaluates on the held-out ones only.

\section{Terminal-Bench Results}
\label{terminalbench}

This appendix gives the Terminal-Bench~4.0 subtask profile (Table~\ref{tab:tb-scorecard}) and the worker's workload (Table~\ref{tab:tb-delegation}).
The lead is GPT-6.1 Sol and the worker GPT-6 Luna, and every method ran twice on the 23 held-out tasks.

\begin{table}[t]
\centering
\small
\setlength{\tabcolsep}{2.5pt}
\renewcommand{\arraystretch}{1.08}
\begin{tabular}{@{}l rr r r ccc@{}}
\toprule
& \multicolumn{2}{c}{Quality} & Saving & & \multicolumn{3}{c}{Tier at $\delta =$} \\
\cmidrule(lr){2-3}\cmidrule(l){6-8}
Subtask & $q_{\Lm}$ & $q_{\Wm}$ & (\%) & $\Risk$ & .10 & .40 & .75 \\
\midrule
Explore environment$^{\ddagger}$ & 7.1 & 6.4 & 85 & .23 & \tK & \tK & \tK \\
Baseline profiling & 6.8 & 3.6 & 69 & .45 & \tK & \tC & \tD \\
Write code \& tests & 6.1 & 3.3 & 89 & .49 & \tK & \tC & \tD \\
Verification & 6.4 & 3.9 & 84 & .42 & \tK & \tC & \tD \\
Debugging & 5.9 & 3.8 & 60 & .46 & \tK & \tC & \tD \\
Finish$^{\ddagger}$ & 5.9 & 3.7 & 87 & .46 & \tK & \tK & \tK \\
\bottomrule
\end{tabular}
\caption{Terminal-Bench subtask quality (0--10), cost saving, risk and tiers from 12 profiling tasks. D = delegate, c = conditional, K = keep. $^{\ddagger}$Exploration and finish stay with the lead.}
\label{tab:tb-scorecard}
\end{table}

\begin{table}[t]
\centering
\small
\setlength{\tabcolsep}{2.5pt}
\renewcommand{\arraystretch}{1.1}
\begin{tabular}{@{}l r rrr@{}}
\toprule
& & \multicolumn{3}{c}{Ours, $\delta =$} \\
\cmidrule(l){3-5}
Measure & Unguided & .10 & .40 & .75 \\
\midrule
Hand-offs & 199 & 0 & 175 & 287 \\
Tasks with worker & 46 & 0 & 46 & 46 \\
Worker turns & 4{,}405 & 0 & 4{,}349 & 7{,}929 \\
Turns/hand-off & 22.1 & -- & 24.9 & 27.6 \\
\midrule
Turn share (\%) & 76.5 & 0.0 & 75.4 & 86.1 \\
Output share (\%) & 84.6 & 0.0 & 84.3 & 93.4 \\
Cost share (\%) & 19.9 & 0.0 & 20.1 & 34.5 \\
\midrule
Median time (min) & 55 & 15 & 39 & 69 \\
\bottomrule
\end{tabular}
\caption{Worker workload per method over two runs of the 23 Terminal-Bench tasks (46 task runs). Shares divide worker totals by lead-plus-worker totals over all task runs. Output tokens include reasoning, commands and reports. At $\delta=0.10$ every subtask is kept, so the lead never delegates. Time is the median wall-clock time per task run; GPT-6.1 Sol alone takes 14 minutes. Turns are model calls.}
\label{tab:tb-delegation}
\end{table}

\section{Second Model Pairs}
\label{app:pair}

We ran a second lead--worker pair on each benchmark, once per application or task: both solo models, and partial routing with schema guidance at $\delta=0.75$.
In both pairs the worker costs a small fraction of what the lead costs and, alone, scores well below it.
These runs check whether the gains of partial routing are specific to the main pairs; they are not meant to show that it helps every pair.

\paragraph{ViBench.}
We ran \vibench with the main Terminal-Bench pair, GPT-6.1 Sol as the lead and GPT-6 Luna as the worker: both solo models on all 15 applications and partial routing on the ten held-out ones.
Both models ran on OpenAI's API at medium reasoning effort, with the main \vibench pair's harness, delegation tool, routing prompt and limits (Table~\ref{tab:limits}), priced at \$2/\$10 for Sol and \$0.10/\$0.50 for Luna per million input/output tokens.
Profiling on the five profiling applications cost about \$13 and induced seven subtasks; at $\delta=0.75$ the lead keeps orientation, planning and delivery and delegates environment setup, backend API, frontend UI and verification.

\begin{figure}[t]
\centering
\begin{tikzpicture}
\tikzset{vb lab/.style={font=\scriptsize, text=black!80, inner sep=1.5pt}}
\begin{axis}[
  width=\columnwidth, height=5cm,
  xmin=0, xmax=0.9, ymin=65, ymax=100,
  xtick={0,0.1,0.2,0.3,0.4,0.5,0.6,0.7,0.8,0.9}, ytick={70,80,90,100},
  xticklabel style={/pgf/number format/.cd, fixed, fixed zerofill, precision=1},
  xlabel={Cost per build (\$)}, ylabel={ViBench score},
  xlabel style={font=\footnotesize, yshift=2pt}, ylabel style={font=\footnotesize, yshift=-4pt},
  ticklabel style={font=\scriptsize},
  grid=major, grid style={black!8}, axis line style={black!45}, tick style={black!45},
  legend cell align=left,
  legend style={at={(0.985,0.03)}, anchor=south east, font=\scriptsize, draw=none, fill=none,
                row sep=-1pt, inner sep=2pt},
]
\addplot[only marks, mark=square*, mark size=2.3pt, soloc] coordinates {(0.816,93.37) (0.036,73.58)};
\addlegendentry{Single model}
\addplot[only marks, mark=diamond*, mark size=3.4pt, mark options={fill=prc!55!black, draw=prc!55!black}] coordinates {(0.2995,88.20)};
\addlegendentry{Partial routing}
\node[vb lab, anchor=east, xshift=-4pt] at (axis cs:0.816,93.37) {GPT-6.1 Sol};
\node[vb lab, anchor=south west, xshift=2pt, yshift=2pt] at (axis cs:0.036,73.58) {GPT-6 Luna};
\node[vb lab, anchor=south, yshift=4pt] at (axis cs:0.2995,88.20) {$\delta{=}0.75$};
\end{axis}
\end{tikzpicture}
\caption{Cost and quality on the ten held-out \vibench applications with GPT-6.1 Sol as the lead and GPT-6 Luna as the worker, one run per application. Partial routing uses schema guidance at $\delta=0.75$. Higher scores and lower costs are better.}
\label{fig:pair}
\end{figure}
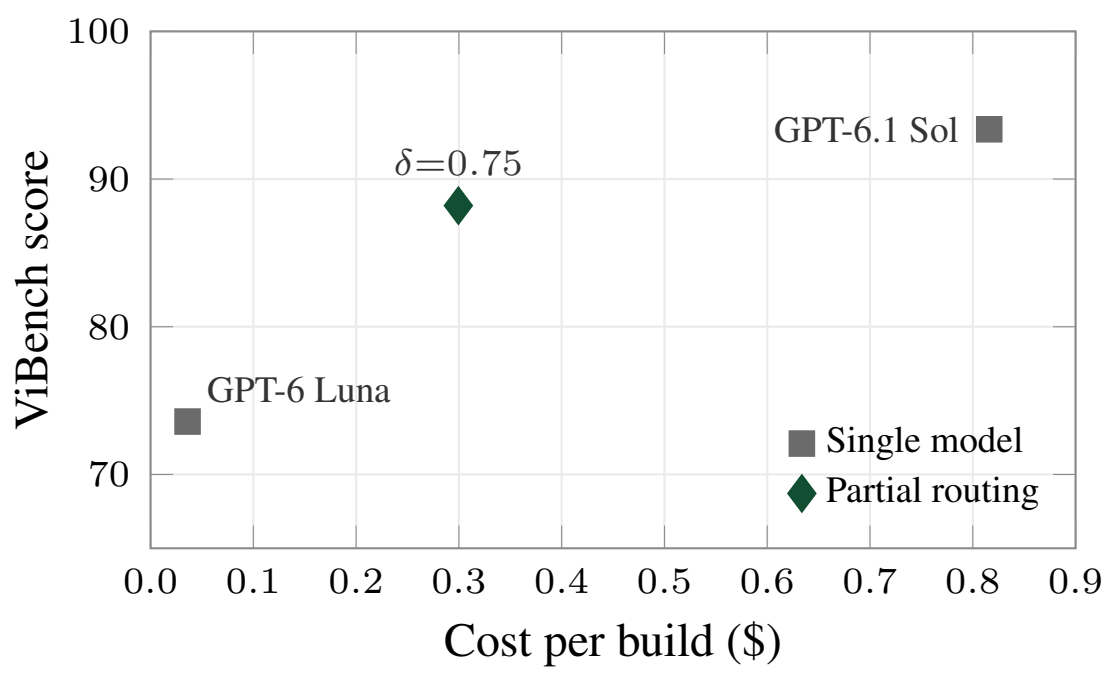

In Figure~\ref{fig:pair}, partial routing scores 88.2 at \$0.30 per build: 5.2 points below Sol alone (93.4 at \$0.82) at 37\% of its cost, and 14.6 points above Luna alone (73.6 at \$0.04).
With a strong, cheap worker, its benefit on this pair is in cost.

\paragraph{Terminal-Bench.}
The second Terminal-Bench pair has Kimi K3.1 Pro as the lead and DeepSeek V4.1 Flash as the worker, priced at \$2/\$10 and \$0.15/\$0.60 per million input/output tokens, respectively.
Kimi K3.1 Pro runs at its default reasoning setting and DeepSeek V4.1 Flash at maximum reasoning effort, alone and as the worker, with the main Terminal-Bench pair's harness, delegation tool, routing prompt and limits (Table~\ref{tab:limits}).
We profiled the pair on the same 12 tasks with the same profiler, $\rho$, $\tau$ and low-confidence rule (Appendix~\ref{app:profile}) for about \$6; the annotator induced seven subtasks, and at $\delta=0.75$ the lead keeps exploration and finishing, delegates implementation, verification, debugging and dependency installation, and decides case by case on probing a reference implementation (low confidence).

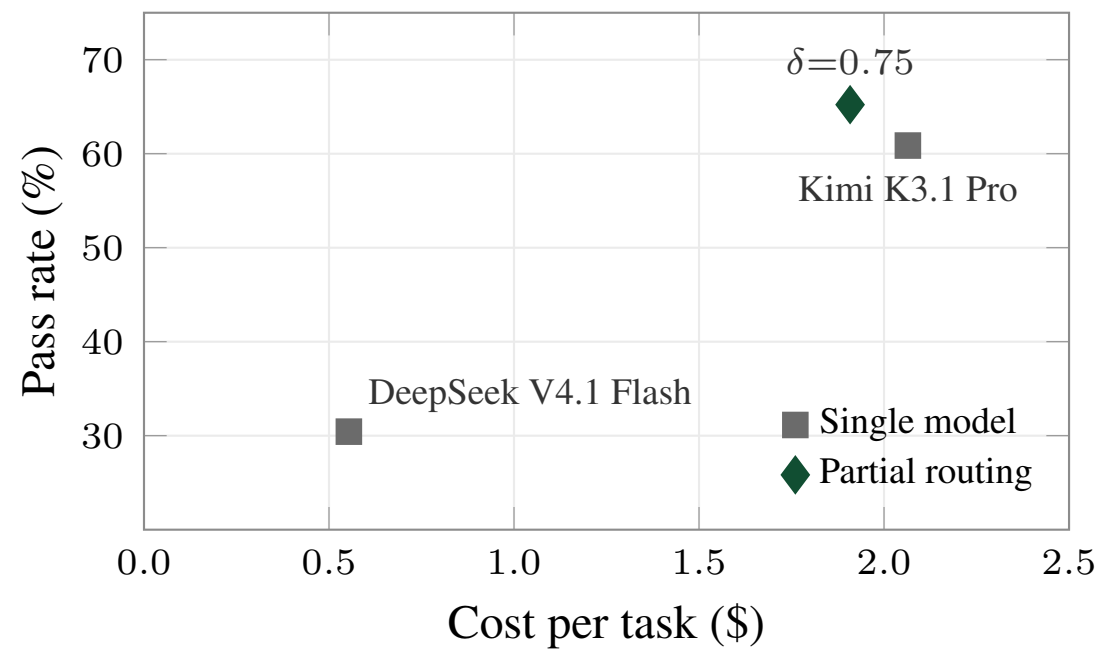
\begin{figure}[t]
\centering
\begin{tikzpicture}
\tikzset{vb lab/.style={font=\scriptsize, text=black!80, inner sep=1.5pt}}
\begin{axis}[
  width=\columnwidth, height=5cm,
  xmin=0, xmax=2.5, ymin=20, ymax=75,
  xtick={0,0.5,1.0,1.5,2.0,2.5}, ytick={30,40,50,60,70},
  xticklabel style={/pgf/number format/.cd, fixed, fixed zerofill, precision=1},
  xlabel={Cost per task (\$)}, ylabel={Pass rate (\%)},
  xlabel style={font=\footnotesize, yshift=2pt}, ylabel style={font=\footnotesize, yshift=-4pt},
  ticklabel style={font=\scriptsize},
  grid=major, grid style={black!8}, axis line style={black!45}, tick style={black!45},
  legend cell align=left,
  legend style={at={(0.985,0.03)}, anchor=south east, font=\scriptsize, draw=none, fill=none,
                row sep=-1pt, inner sep=2pt},
]
\addplot[only marks, mark=square*, mark size=2.3pt, soloc] coordinates {(2.0645,60.87) (0.5537,30.43)};
\addlegendentry{Single model}
\addplot[only marks, mark=diamond*, mark size=3.4pt, mark options={fill=prc!55!black, draw=prc!55!black}] coordinates {(1.9083,65.22)};
\addlegendentry{Partial routing}
\node[vb lab, anchor=north, yshift=-4pt] at (axis cs:2.0645,60.87) {Kimi K3.1 Pro};
\node[vb lab, anchor=south west, xshift=2pt, yshift=2pt] at (axis cs:0.5537,30.43) {DeepSeek V4.1 Flash};
\node[vb lab, anchor=south, yshift=4pt] at (axis cs:1.9083,65.22) {$\delta{=}0.75$};
\end{axis}
\end{tikzpicture}
\caption{Cost and pass rate on the 23 held-out Terminal-Bench tasks with Kimi K3.1 Pro as the lead and DeepSeek V4.1 Flash as the worker, one run per task. Partial routing uses schema guidance at $\delta=0.75$. Higher pass rates and lower costs are better.}
\label{fig:pair-tb}
\end{figure}

On the 23 held-out tasks, partial routing passes 15 tasks (65.2\%) at \$1.91 per task, against 14 (60.9\%) at \$2.06 for Kimi alone and 7 (30.4\%) at \$0.55 for DeepSeek alone (Figure~\ref{fig:pair-tb}).
On this pair its benefit is a higher pass rate than the lead alone at 8\% lower cost.
\end{document}